\pdfoutput=1

\documentclass[11pt]{article}

\usepackage[]{latex/acl}

\usepackage{times}
\usepackage{latexsym}

\usepackage[T1]{fontenc}

\usepackage[utf8]{inputenc}

\usepackage{url}
\usepackage{nicefrac}
\usepackage{booktabs}
\usepackage{amsfonts}
\usepackage{bold-extra}
\usepackage{multirow}
\usepackage{multicol}
\usepackage{xspace}
\usepackage{xcolor}
\usepackage{makecell}
\usepackage{array}
\usepackage{url}
\usepackage{amsmath}
\usepackage{tabularx}
\usepackage{dcolumn,caption}
\usepackage{arydshln}
\usepackage{tikz}
\usepackage{pgfplots}
\usepackage{framed}
\usepackage{pifont}
\usepackage{bbm}
\usepackage{enumitem}
\usepackage{subcaption}
\usepackage{arydshln}
\usepackage{amssymb}
\usepackage{tcolorbox}
\usepackage{adjustbox}
\usepackage{pifont}
\usepackage{mdframed}
\usepackage{subcaption}
\usepackage{verbatim}
\usepackage{fancyvrb}
\usepackage{wrapfig}
\usepackage{graphicx}
\usepackage{subcaption}
\usepackage{booktabs}
\usepackage{fancyvrb}
\usepackage{xcolor}
\usepackage{tcolorbox}
\usepackage{color}
\usepackage{etoc}
\usepackage{makecell}
\usepackage{multicol}
\usepackage{microtype}
\usepackage{inconsolata}

\usepackage{CJKutf8}

\usepackage[title]{appendix}
\newcommand{\eval}[0]{\textsc{S3Eval}\xspace}

\newcommand{\icon}{\raisebox{-1pt}{\includegraphics[width=1.2em]{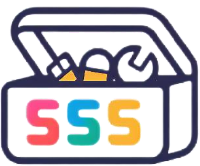}}\xspace}

\title{\icon S3Eval: A Synthetic, Scalable, Systematic Evaluation Suite for \\Large Language Models}

\author{
\makecell{Fangyu Lei\,$^{1,2}$\thanks{~~Equal Contributions.}~~,~Qian Liu\,$^{3*}$,~Yiming Huang\,$^{1*}$,\\
Shizhu He\,$^{1,2}$,~Jun Zhao\,$^{1,2}$,~Kang Liu\,$^{1,2}$}
\vspace{2mm}
\\
\centerline{$^1$Institute of Automation, CAS~~$^2$University of Chinese Academy of Sciences~~$^3$Sea AI Lab}\\
\texttt{leifangyu2022@ia.ac.cn}~~~
\texttt{liuqian@sea.com}~~~
\texttt{kliu@nlpr.ia.ac.cn}
}

\begin{document}
\maketitle
\begin{abstract}
The rapid development of Large Language Models (LLMs) has led to great strides in model capabilities like long-context understanding and reasoning.
However, as LLMs are able to process longer contexts, it becomes more challenging to evaluate whether they have acquired certain capabilities, since the length of text (e.g., 200K tokens) they can process far exceeds what humans can reliably assess in a reasonable duration.
In this paper, we propose using complex synthetic tasks as a proxy evaluation method, and present \eval, a \textbf{S}ynthetic, \textbf{S}calable, \textbf{S}ystematic evaluation suite for LLMs evaluation.
The synthetic nature of \eval provides users full control over the dataset, allowing them to systematically probe LLM capabilities by scaling text length and varying task difficulty across diverse scenarios.
The strong correlation between \eval and real-world benchmarks demonstrates the soundness of using \eval for evaluation of LLMs.
\eval provides a flexible and infinite long-context data generation method. We have generated a comprehensive dataset called \eval-Standard, and experimental results have shown that it poses significant challenges for all existing LLMs.
Our code is available at \url{https://github.com/lfy79001/S3Eval}.
\end{abstract}

\section{Introduction}

Large Language Models (LLMs) have greatly propelled significant advancements in Natural Language Processing (NLP), such as OpenAI GPT~\citep{brown2020language}, Llama~\citep{touvron2023llama,touvron2023llama2}, StarCoder~\citep{li2023starcoder}, and others. These models perform well in many NLP tasks and claim to have made progress in advanced capabilities such as reasoning, long-context understanding, and so on. However, existing benchmarks~\citep{chang2023survey} often fail when it comes to evaluating extremely long-context LLMs or analysing the controllable characteristics and limitations of LLMs.

\begin{figure}[tb]
  \centering
  \includegraphics[width=0.48\textwidth]{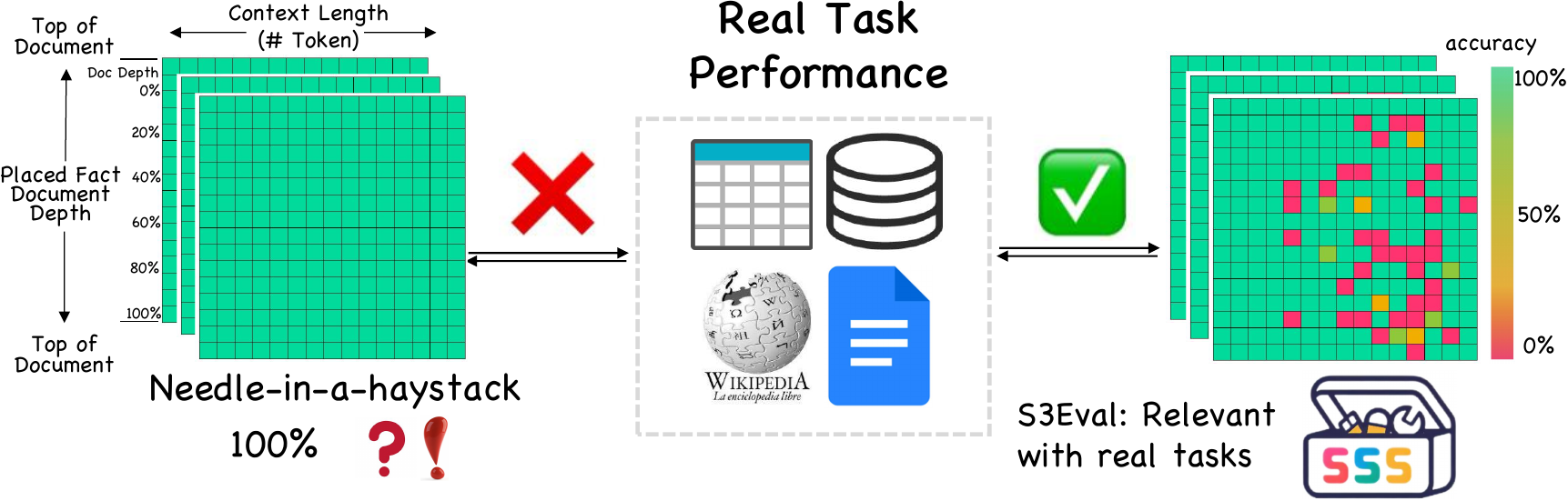}  \caption{\textit{Needle-in-A-HayStack} cannot demonstrate the performance of the model under real tasks, but \eval can. Compared with \textit{needle-in-a-haystack}, \eval is more relevant to real benchmarks and more difficult.}
    \label{fig:needle}
\end{figure}
 
For long-context understanding, previous work has often evaluated LLMs using the scope of language modeling metrics (i.e., perplexity)~\citep{sun2021long, peng2023yarn} or the performance on simple artificial tasks~\citep{li-roth-2002-learning, berant2013semantic, mohtashami2023landmark}. 
There is a widely used evaluation method known as \textit{Needle-in-a-Haystack}~\citep{Kamradt2023}, as shown in Figure~\ref{fig:needle}. In this method, a vital piece of information is concealed within a lengthy document, resembling a haystack, and the model's objective is to locate and retrieve this hidden key information.
However, these evaluation tasks tend to lack complexity and are narrowly focused on simple comprehension, which is misaligned with the sophistication required for real-world downstream applications.

While recent work has made great progress on building evaluation benchmarks at longer context lengths with real-world use cases (e.g, question answering)~\citep{bai2023longbench,an2023eval}, these manually annotated datasets often lack the scale and diversity to thoroughly assess performance on extended context lengths.
For example, existing benchmarks struggle to effectively evaluate LLMs that claim an ability to process contexts up to 100K tokens, due to the limited capacity of human annotation for very long text.
Developing more scalable and diverse evaluation datasets, potentially leveraging automated supervision, remains an open challenge.

For reasoning analysis~\citep{math2021, chen2021evaluating, suzgun2022challenging, zhong2023agieval}, conducting both qualitative and quantitative analysis of answers and reasoning processes provides important insights.
However, existing benchmarks lack the ability to precisely control the distribution of the dataset, limiting their utility for in-depth research analysis.
In other words, the nature of these benchmarks makes it challenging for developers to identify the specific weaknesses of their LLMs.
More configurable and granular benchmarks are needed to enable detailed analysis of model performance.
In addition, these benchmarks often draw their evaluation data from NLP tasks that have been extensively studied and are likely to be used in the training corpus of LLMs.
The potential data leakage makes the evaluation less convincing.

\begin{figure*}[t]
    \centering
	\includegraphics[width=\textwidth]{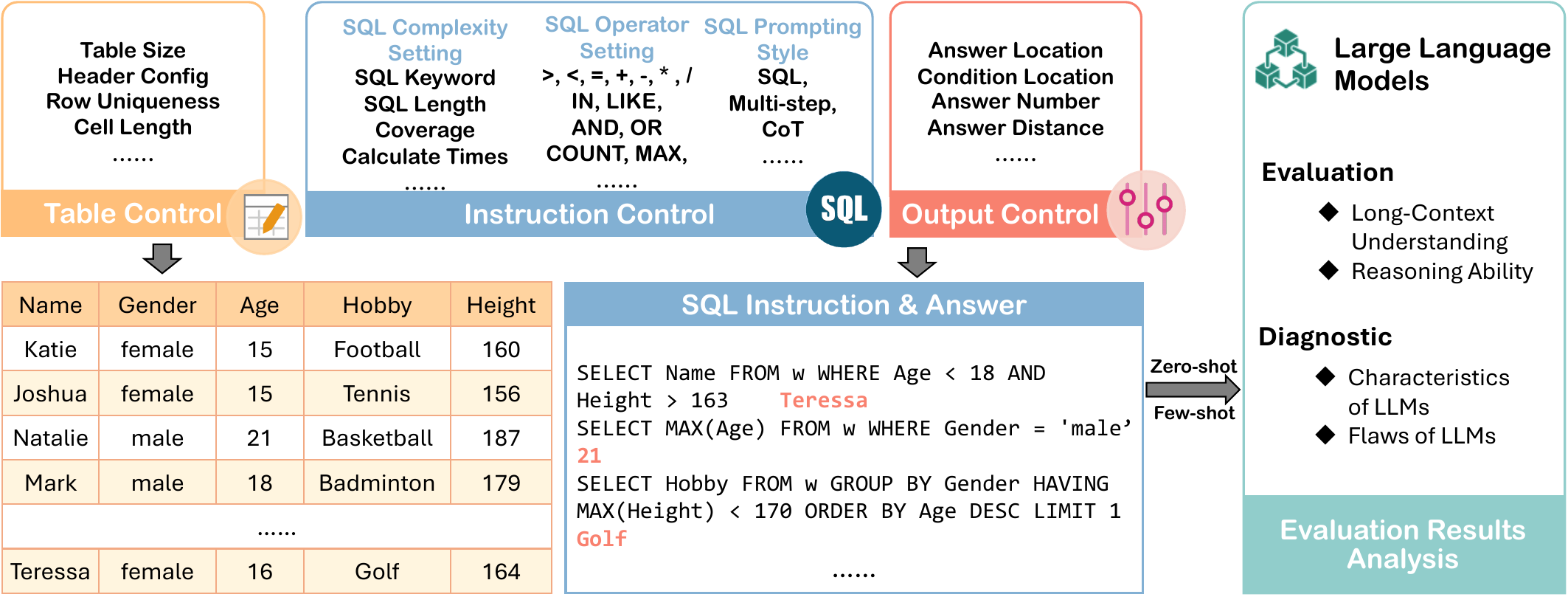}
	\caption{The illustration demonstrates the \eval pipeline, where the capabilities of LLMs are assessed by evaluating their ability to execute SQL queries over randomly generated tables.}
	\label{fig:pipeline}
\end{figure*}

In this paper, we propose a new evaluation suite called \eval, which addresses the aforementioned issues by using a complex synthetic task - SQL execution - as a proxy for the performance of LLMs on realistic reasoning tasks.
As shown in Figure~\ref{fig:pipeline}, inspired by the work of \textsc{TaPEx}~\citep{liu2021tapex}, \eval is based on the SQL execution task.
Specially, given a randomly generated table and a random SQL query, \eval evaluates whether LLMs can return the correct execution results.
\eval has three notable characteristics: (1) It is synthetic, with no table or SQL query present in the LLM training corpus. The tasks use complex, grammatically correct SQL syntax, making them very challenging. 
(2) It is scalable, allowing users to customize the benchmark to any length and difficulty.
(3) It is systematic, containing diverse reasoning types and operations. This enables comprehensive evaluation of LLM capabilities.

With these powerful features, developers can extend the context to really long lengths and generate meaningful SQL statements using \eval.
We conducted comprehensive multi-perspective experiments on several popular LLMs using \eval.
Experimental results demonstrated that the performance of LLMs on \eval aligns closely with their performance on mainstream LLM benchmarks.
While LLMs have shown impressive capabilities, our work reveals limitations in their ability to leverage long contexts, since we observe performance degradation of almost all LLMs in long-context settings.
By carefully studying experimental results, we can work to pinpoint situations where LLMs tend to fail and summarize valuable insights. 

In the era of rapid LLM development, the most significant contribution of \eval lies in its effectiveness as a method for long-context evaluation. Capable of generating evaluation data of infinite length, it ensures that assessments are not only reasonable but also sufficiently challenging.

\section{Synthetic: Suite and Benchmark}
In this section, we introduce the details of the \eval evaluation suite~(as shown in Figure~\ref{fig:pipeline}) and the new benchmark we proposed. 

\begin{table*}[hbtp]
\centering
\small
\setlength{\tabcolsep}{2.0mm}
\begin{tabular}{lll}
\toprule
\multicolumn{2}{c}{\textbf{Configuration}}                      & \multicolumn{1}{c}{\textbf{Description}}            \\ \midrule
\multirow{5}{*}{Table Control}      & \# of Rows &  The number of rows in the generated tables         \\
                        & \# of Columns &  The number of columns in the generated tables         \\
                            & Header Type Ratio    & The proportion of table column types that are TEXT, INT, DATE        \\
                            & Cell Uniqueness & The proportion of duplicate cells in each column       \\
                            & String / Int Length  & The string length or numeric range of cell values         \\ \midrule
\multirow{8}{*}{\shortstack[c]{Instruction\\Control}} & SQL Keywords         & SELECT,  WHERE, GROUP BY, HAVING, ORDER BY \\
                            & SQL Length           & The number of tokens after SQL split by space  \\
                            & Column Coverage      & The ratio of columns involved in SQL execution to total columns.  \\
                            & Row Coverage         & The ratio of rows involved in SQL execution to total rows   \\
                            & Calculate Times      & The number of SQL numerical calculations.  \\
                            & Filter Times         & The number of SQL filtering operations.      \\
                            & Aggregator           & COUNT, MAX, MIN, SUM, AVG                  \\
                            & Filter Operator      & \textgreater{}, \textless{}, =, IN, LIKE   \\  \midrule
\multirow{3}{*}{Output Control} & Answer Location      & The location of SQL answers in the input table     \\ 
                               & \# of Answer Cells       & The number of selected cells in the answer   \\
                               & Answer Length      & The total number of tokens in the answer      \\      \bottomrule
\end{tabular}
\caption{Our \eval method allows users to customize configuration settings and provides descriptions for each parameter that can be adjusted. More configurations can be found in Appendix~\ref{app:setting_des}.}
\label{tab:configure}
\end{table*}
\begin{table*}[t]
\centering
\small
\begin{tabular}{lclll}
\toprule
Model                & Context Length & Short-Context & Long-Context & Total \\ \midrule
GPT-4-32K            &  32768  &     68.4\% &       43.0\%    &     54.8\%   \\
GPT-3.5-Turbo        &  16384  &     39.9\%       &       16.2\%     &    27.0\% \\
Code Llama (70B)     &  16384  &     33.9\%        &     8.9\%      &   20.3\%    \\
LLaMA-2 (70B)        &  4096   &     30.0\%          &      8.8\%       &    18.4\%   \\
LLaMA-2 (13B)        &  4096  &      21.7\%         &       4.6\%     &     12.4\%  \\
LLaMA-2 (7B)         &  4096   &     20.8\%          &      4.4\%      &   11.9\%  \\
Gemma (7B)           &  8192   &     28.9\%        &       8.6\%       &   17.9\%   \\
Qwen 1.5 (14B)       &  32768   &    33.7\%        &    14.4\%        &     23.2\%  \\
Qwen 1.5 (7B)        &  32768  &     26.5\%      &      8.0\%       &    16.5\%   \\
Qwen 1.5 (4B)        &  32768  &     22.8\%      &      5.5\%       &    13.4\%   \\
Mixtral-8x7B (46.7B) &  32768  &     31.5\%       &     11.1\%      &   20.4\%  \\ 
Mistral-Instruct-v0.2 (7B) & 32768 &  28.7\%       &     10.6\%      &   18.9\%  \\

\bottomrule
\end{tabular}
\caption{Experimental results on S3Eval-Standard.  ``Total'' denotes the overall score, ``Short-Context'' refers to the model's performance on contexts shorter than 4K in length, and ``Long-Context'' indicates the model's performance on contexts ranging from 4K to 40K in length.}
\label{tab:new_main_results}
\end{table*}

\subsection{Suite Construction}
\label{sec:method}

\paragraph{Task Formulation}

Following previous work~\citep{liu2021tapex}, each example in \eval generally contain an SQL query and a (semi-)structured table $T$ as the input.
Each table $T$ consists of $M$ rows $\left \{ r_i \right \}_{i=1}^{M}$, in which each row $r_i$ contains $N$ cell values $\left \{ c_{\left \langle i,j \right \rangle }  \right \} _{i=1}^{M} $. Each cell $c_{\left \langle i,j \right \rangle }$ corresponds to a table header $h_{j}$. Each SQL query consists of $K$ tokens as $x=x_1,x_2,\cdots,x_K$. Each token $x_i$ originates from SQL keywords, table schema, or table cells. Each multi-step instruction is transformed from SQL query. The task prompts LLM to obtain the execution result $A$ of the SQL on the table $T$. Our main focus is on analyzing the accuracy of LLM in executing SQL queries.

\paragraph{Random Table Generation}

All tables in \eval{} are randomly generated and do not contain any real data or overlap with existing public tables. The tables have $M$ rows and $N$ columns, with adjustable parameters $M$ and $N$. The column headers are sampled from English nouns~\citep{bird2006nltk}, falling into three types: TEXT, INT, and DATE. INT columns contain random integers from 1 to 1000, which is an adjustable range. DATE columns have values in year-month-day format. TEXT columns have random strings of length 5 to 12 characters, which is also adjustable.
To simulate real-world data where the same value may recur in a column frequently, the data generator includes a parameter to set the probability of duplicating values within a specific column.


\paragraph{Random SQL Generation}

The SQL language includes a variety of statements to query and manage data. \eval use context free grammar to generate a specific number of examples with controllable attributes.
As Table~\ref{tab:configure} shows, the \eval tool allows configuring several parameters of generated SQL statements, including nesting depth, keywords used, length, coverage of SQL features, computational complexity, and more. 
For example, \textit{calculate times} can be modified to control the complexity of numerical reasoning for each dataset. Except these configures, users can also manually write the specified SQL template to generate fine-grained evaluation data~(Appendix~\ref{app:template}).


\paragraph{Evaluation Methods}

\eval{} includes both zero-shot and few-shot prompting methods. For each few-shot setting, all examples share one table. N-shot is formalized as $\mathrm{INPUT = [T;S_1;A_1;...;S_{n+1}]}$. For the input format of table $T$, we designed several alternative ways, including markdown, flatten, tapex-style, etc.

To evaluate the performance of LLMs, we use Exact Match (EM) as the evaluation metric.
Details are shown in Appendix~\ref{app:table_input_format}.

\subsection{\eval-Standard Benchmark}
We generate a highly diverse dataset called \eval-Standard covering lengths ranging from 2K to 40K, with various difficulty levels of reasoning types, which comprises all templates and operations included in \eval, making it the most complex and diverse dataset available. We utilize this version of the dataset as the official benchmarking data for \eval benchmark. It can effectively evaluate LLMs in completing tasks under both short-context and long-context scenarios. We evaluate popular commercial LLMs and open-source LLMs on \eval-Standard, and the experimental results are shown in Table~\ref{tab:new_main_results}. In theory, we can measure the performance of LLMs with unlimited context length here.

\section{Correlation with Realistic Benchmark}

In this section, we describe the details of synthesizing the evaluation data~(Section~\ref{sec:method}) and verify the correlation between our synthetic suite \eval and real-world benchmark results.

\subsection{Experimentual Settings}

\eval can flexibly generate different evaluation data. To validate the rationality of \eval, we conducted correlation experiments and generated two sets of data with different difficulty levels for experimentation.
\textbf{Easy} is the simplest data that \eval can generate and is used to evaluate LLM’s ability to understand the most basic instructions. It contains only one template, ``SELECT \textless{}col1\textgreater{} WHERE \textless{}col2\textgreater{} \textless{}op\textgreater{} \textless{}value\textgreater{}''.
\textbf{General} is a more difficult setting, containing extensive SQL syntax, and its generating setting is described in Appendix~\ref{app:general_setting}.
All experiments were run for $3$ times, using 1000 randomly generated queries per trial, with tables of 15 rows and 8 columns and an average of 1200 tokens per input. Details on the LLMs are provided in Appendix~\ref{app:llm_details}.

Considering \textit{SQL execution} is a difficult task, some models may have a poor understanding of symbolic language, which makes it difficult to execute SQL, so we propose an alternative task \textit{SQL multi-step task} to remove this potential bias. Specifically, it converts an SQL query into a multi-step table operation instruction as shown in Appendix~\ref{app:multi_step}. SQL has a fixed execution flow for the query statement: FROM $\rightarrow$ ON $\rightarrow$ JOIN $\rightarrow$ WHERE $\rightarrow$ GROUP BY $\rightarrow$ HAVING $\rightarrow$ SELECT $\rightarrow$ ORDER BY $\rightarrow$ LIMIT. This is not consistent with the order in which it is written. With this processing, it can also generate chain-of-thought prompting data.

\begin{figure*}[t]
  \centering
    \begin{minipage}[b]{0.45\textwidth}
        \centering
    \includegraphics[width=0.9\textwidth]{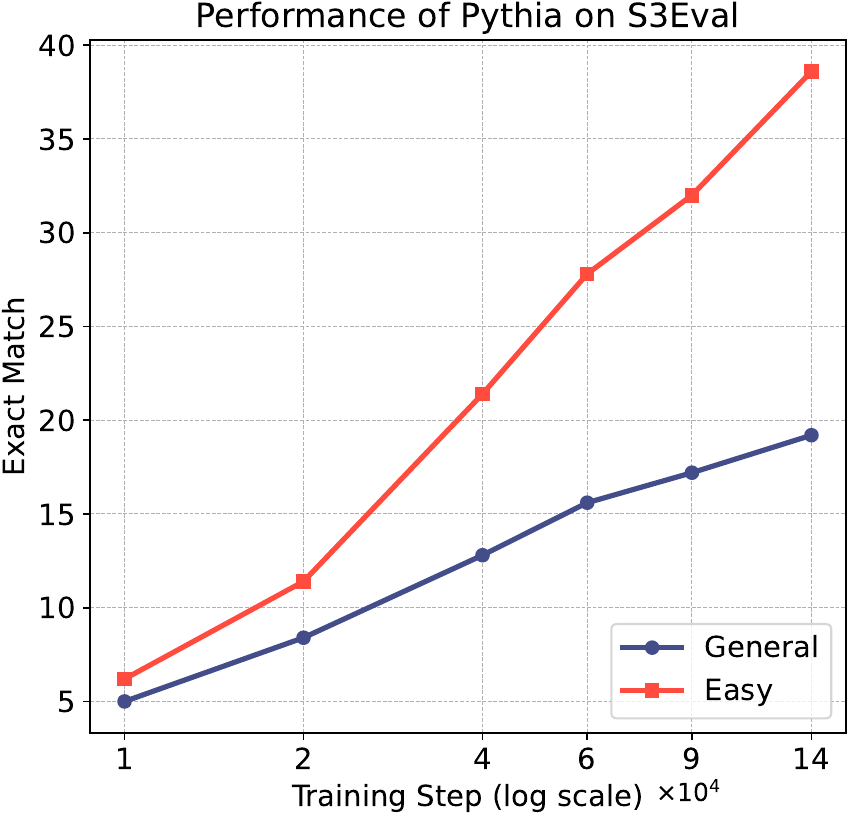}
    \caption{The performance of Pythia-12B on \eval was evaluated across different training steps.}
    \label{fig:pythia}
  \end{minipage}
  \hfill
  \begin{minipage}[b]{0.45\textwidth}
    \centering
    \includegraphics[width=1.0\textwidth]{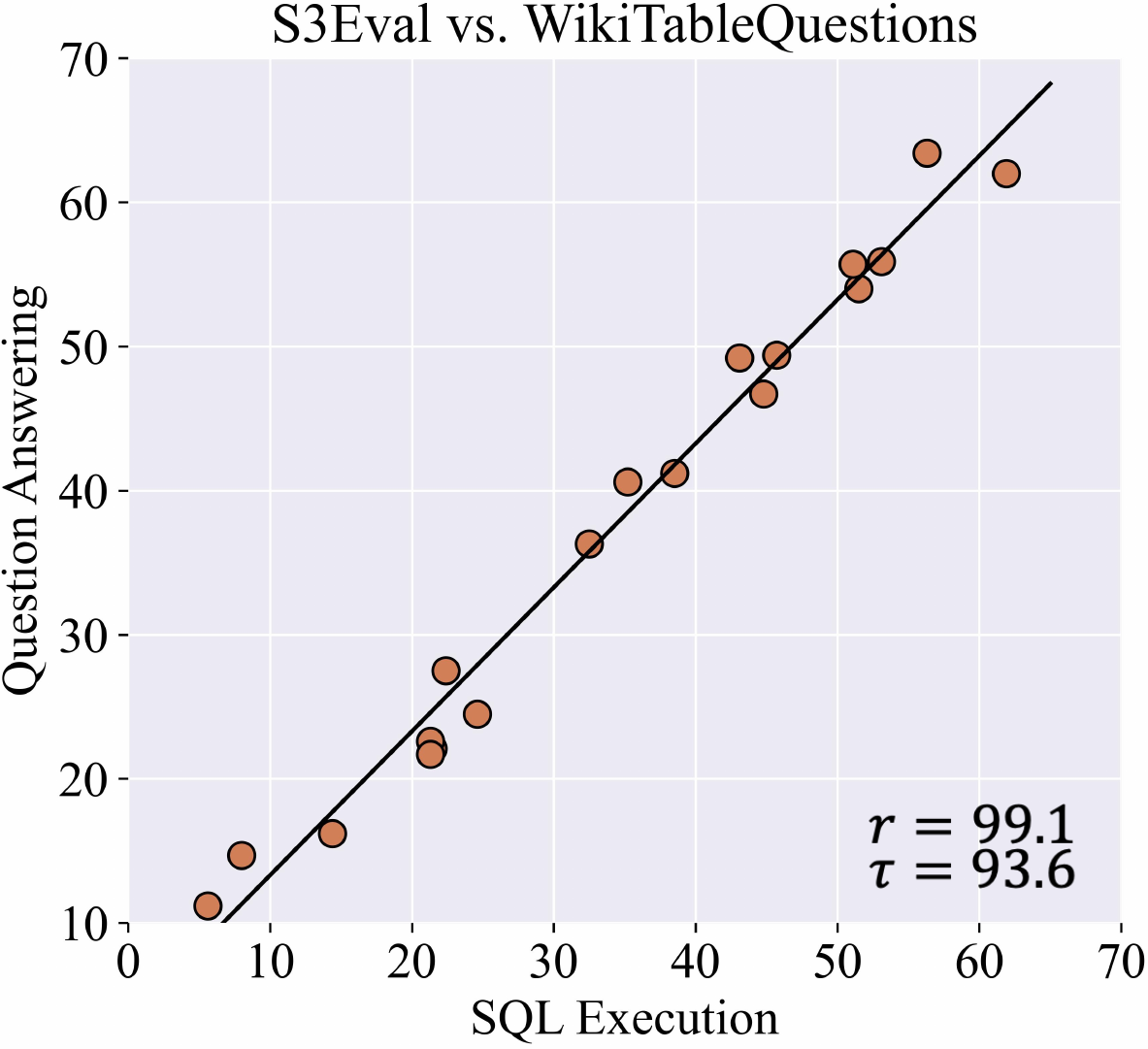}
    \caption{The performance of different LLMs on \eval and WikiTableQuestions.}
    \label{fig:wtq_alignment}
  \end{minipage}
\end{figure*}

\subsection{Scaling Law}

Previous work~\citep{scaling_law,hoffmann2022training} shows a positive correlation between the cross-entropy loss of LLMs and the amount of computing resources used for training, as described by the empirical scaling law.
To verify whether the scaling law holds for our \eval, we employ a set of checkpoints of Pythia-12B~\citep{biderman2023pythia} that are open-sourced at different training steps, corresponding to different amounts of compute.
We observe a consistent pattern as illustrated in Figure~\ref{fig:pythia}: the scores show a smooth progression of improvement that aligns with the scaling law with increasing the training steps.
The steady, incremental performance gains over time, lacking any spikes, demonstrate \eval's reliability as a evaluation suite.
Overall, these experimental results confirm the scaling law's accuracy in forecasting model gains during training across diverse evaluation settings.

\subsection{Benchmark Performance}

In the above, we validated that the LLMs also exhibits the scaling law observed in NL on the \eval suite. A natural question that arises is whether its performance on \eval is correlated with the performance on real-world, NL benchmarks.
To examine the hypothesis, we first compare the performance of different LLMs on \eval and on WikiTableQuestions~\citep{pasupat2015compositional}, a table question answering dataset consisting of questions and answers.
It is worth noting that to align the difficulty, we use the SQL queries from WikiTableQuestions~\citep{shi2020potential} as our \eval evaluation set.

To systematically compare the performance, following previous work~\citep{liu2021question}, 
we consider two correlation measures: the Pearson correlation coefficient ($r$), which evaluates the linear relationship between model scores on the two benchmarks, and the Kendall rank correlation coefficient ($\tau$), which assesses whether the relative ranking of models is consistent across the benchmarks.
The strong correlation between LLMs' performance on the SQL execution task and the table question answering task, as evidenced by the high $r$ (e.g., 99.1) and high $\tau$ (e.g., 93.6) in Figure~\ref{fig:wtq_alignment}.

\begin{figure*}[tb]
    \centering    
    \subcaptionbox{Performance of large language models\label{fig:bbh_general}}{
    \includegraphics[width=0.43\textwidth]{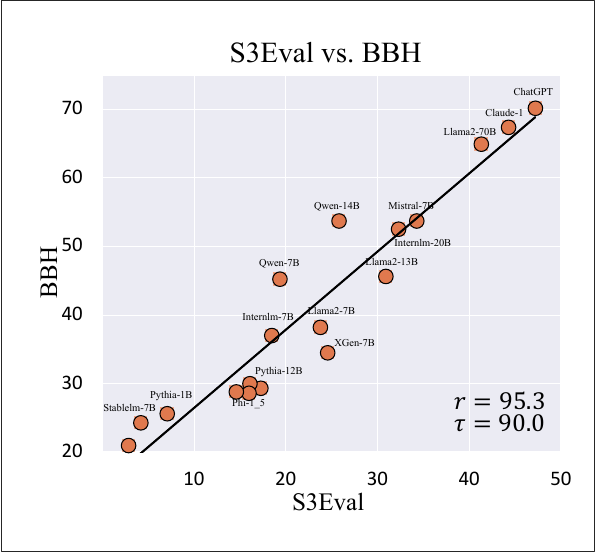}
    }
    \hfill
    \subcaptionbox{Performance of code large language models\label{fig:code_general_human}}{
    \includegraphics[width=0.43\textwidth]{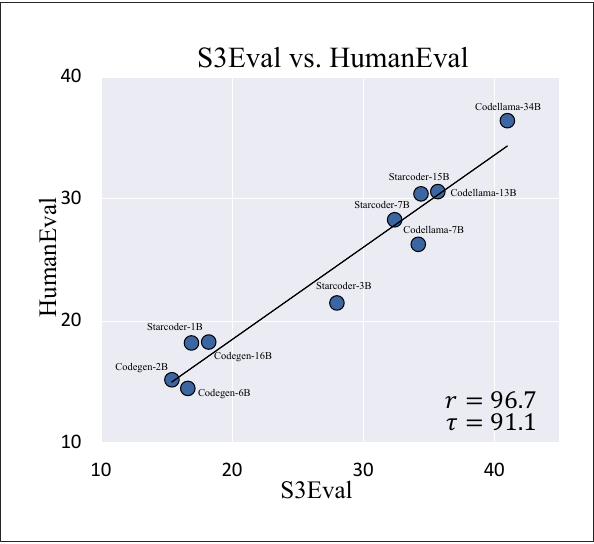}
    }
    \caption{Each point in the scatterplot represents the LLM performance on the benchmarks corresponding to the horizontal and vertical coordinates. The black straight line is the trend line. The larger the values of $r$ and $\tau$, the higher the correlation between the two benchmarks. We consider $\tau$ \textgreater{} 0.8 to be high concurrence.}
    \label{fig:general_alignment}
\end{figure*}

Although \eval has shown significant correlation with WikiTableQuestions, the fact that they are both tasks on tables may cause one to question whether \eval can serve as a proxy task to evaluate LLMs' capabilities on generic reasoning tasks.
Therefore, we also compare the performance on \eval with the results of generic popular benchmarks like BBH~\citep{suzgun2022challenging} and HumanEval~\citep{chen2021evaluating}.
The results depicted in Figure~\ref{fig:bbh_general} demonstrate a strong correlation between LLM performance on \eval and the BBH benchmark, with BBH performance obtained from the OpenCompass platform using few-shot chain-of-thought prompting~\citep{2023opencompass}.
Similarly, Figure~\ref{fig:code_general_human} illustrates the correlation between \eval performance and pass@1 scores on HumanEval~\citep{chen2021codex} for code LLMs.
The results demonstrate that \eval serves as a robust proxy task for assessing the reasoning capabilities of LLMs on realistic benchmarks.
Concrete experimental results are provided in Table~\ref{tab:mail_results}.

\begin{figure*}[htbp]
  \centering
\begin{minipage}[b]{0.48\textwidth}
    \centering
    \includegraphics[width=\textwidth]{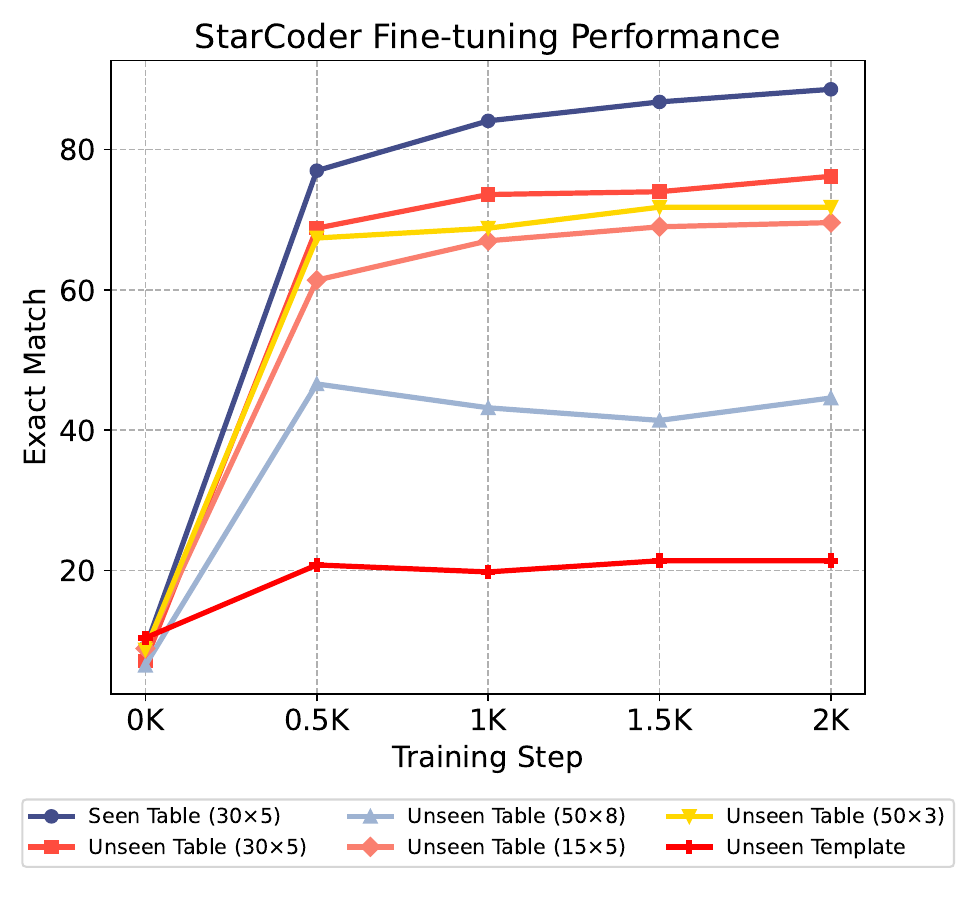}
    \caption{SQL execution training experiments on \eval.}
    \label{fig:starcoder}
  \end{minipage}
\hfill
  \begin{minipage}[b]{0.495\textwidth}
    \centering
    \includegraphics[width=\textwidth]{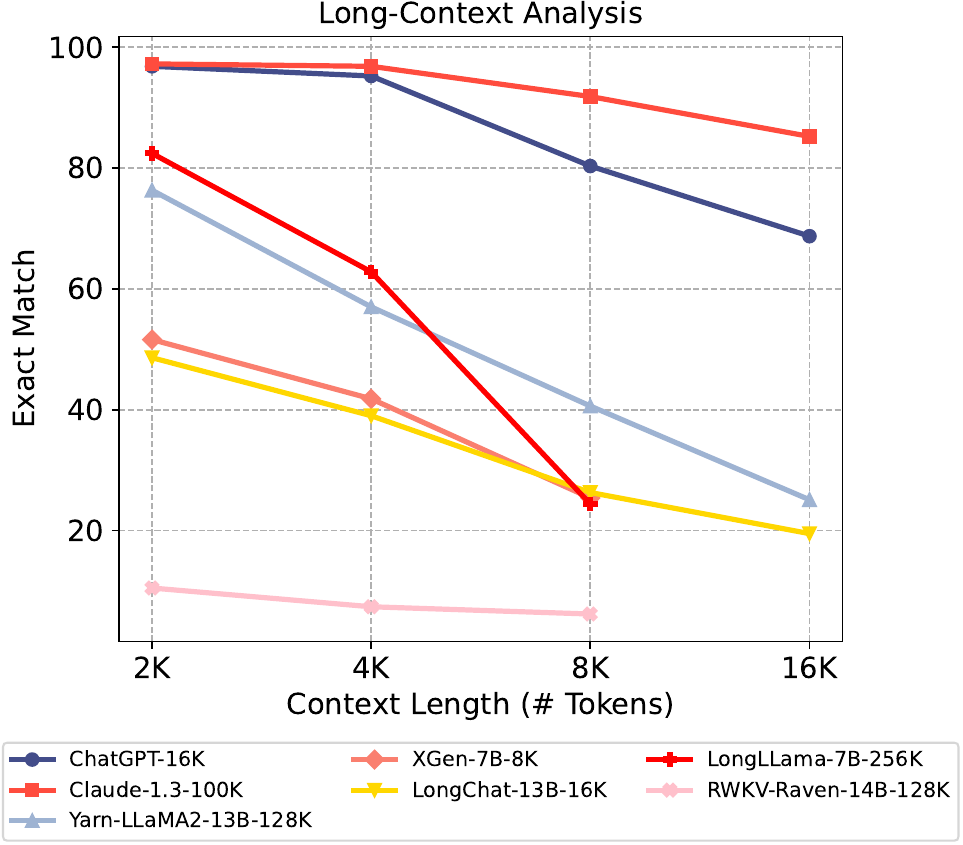}
    \caption{Experiment results of different LLMs on different context lengths.}
    \label{fig:long_context}
  \end{minipage}
\end{figure*}

\section{Scalable: Unlimited Evaluation Resources}
\label{Scalable}

\eval provides a unique capability to generate infinite number of examples~(Section~\ref{sec:scalable_number}) with infinite length~(Section~\ref{sec:scalable_length}).

\subsection{Scalable Number of Evaluation Examples}
\label{sec:scalable_number}

The strength of \eval is its ability to generate unlimited number of examples for evaluation. This stems from two key design choices in \eval: (1) the synthetic table size can be scaled to different number of rows and columns, and (2) the table cells are synthesized from randomly generated strings. Combined with the provided large library of SQL query templates, these features enable the creation of a near-infinite set of unique evaluation examples.
This kind of capacity enables the continuous creation of novel examples unseen during training, which helps safeguard test data integrity by preventing leakage of the evaluation set into the training corpus.

However, the absence of data leakage does not necessarily mean that \eval's performance always represents the model's out-of-distribution generalization ability. It is because the model may perform well on \eval via domain-specific training on the SQL execution task, rather than acquiring more general abilities.
To investigate whether LLMs can ``hack'' \eval via domain-specific training, we fine-tuned StarCoder-1B~\citep{li2023starcoder}, which is not able to solve SQL execution tasks, on a randomly generated dataset of one million examples. 
The performance of the fine-tuned StarCoder-1B is illustrated in Figure~\ref{fig:starcoder}, where it is evaluated on three types of test datasets: \texttt{Seen Table} (same tables as training), \texttt{Unseen Table} (new tables in same format as training tables), and \texttt{Unseen Templates} (new SQL query templates).
For the unseen table setting, we explore different table shapes, where $(x\times y)$ means the table consists of $x$ rows and $y$ columns.

The experimental results demonstrate that for Unseen Tables with different shapes, regardless of their size, the performance of the fine-tuned StarCoder experiences a substantial decline compared to Seen Tables. Likewise, when faced with Unseen Templates, the performance of the fine-tuned StarCoder exhibits a significant drop. 
The results indicate that even if LLMs have been heavily trained on SQL execution tasks, their out-of-distribution performance can still be accurately evaluated by using novel SQL templates. These new SQL templates can be easily generated thanks to the vast grammar of SQL queries. Additionally, evaluating LLMs on larger tables that they were not trained on can also reveal part of their out-of-distribution capabilities.

\subsection{Scalable Length of Evaluation Examples}
\label{sec:scalable_length}

One advantage of \eval is its scalability and adjustable context length per example.
The flexibility allows \eval to rigorously evaluate LLMs that claim capability with long contexts.
To clearly expose limitations of current LLMs, we intentionally chose the \textbf{Easy} setting in \eval to evaluate their performance. Specifically, we establish table configurations with approximately 2K, 4K, 8K, and 16K tokens, by using different numbers of rows and fixing the number of columns.
We generate a dataset consisting of 500 samples for each evaluation setting.
The experimental results on up to 16K context length are plotted in Figure~\ref{fig:long_context}.
As observed, the performance of almost all LLMs, significantly decreases as the context length increases.
Of all the models, Claude-1.3-100K is the only one that maintains a relatively strong performance trend.
Detailed results can be found in Appendix~\ref{app:long_context}. 

As illustrated in Table~\ref{tab:new_main_results}, \eval poses significant challenges for models even when the context window is extended to 32K levels. This difficulty arises from \eval being rooted in real-world tasks, enabling it to generate evaluation data of infinite length and ensure the tasks are both reasonable and demanding. Looking ahead, as models progress to the 200K level, \eval will likewise be poised to furnish effective evaluation data.

\begin{figure}[tb]
    \subcaptionbox{ChatGPT\label{fig:gpt_dot2}}{
        \includegraphics[width=0.4\textwidth]{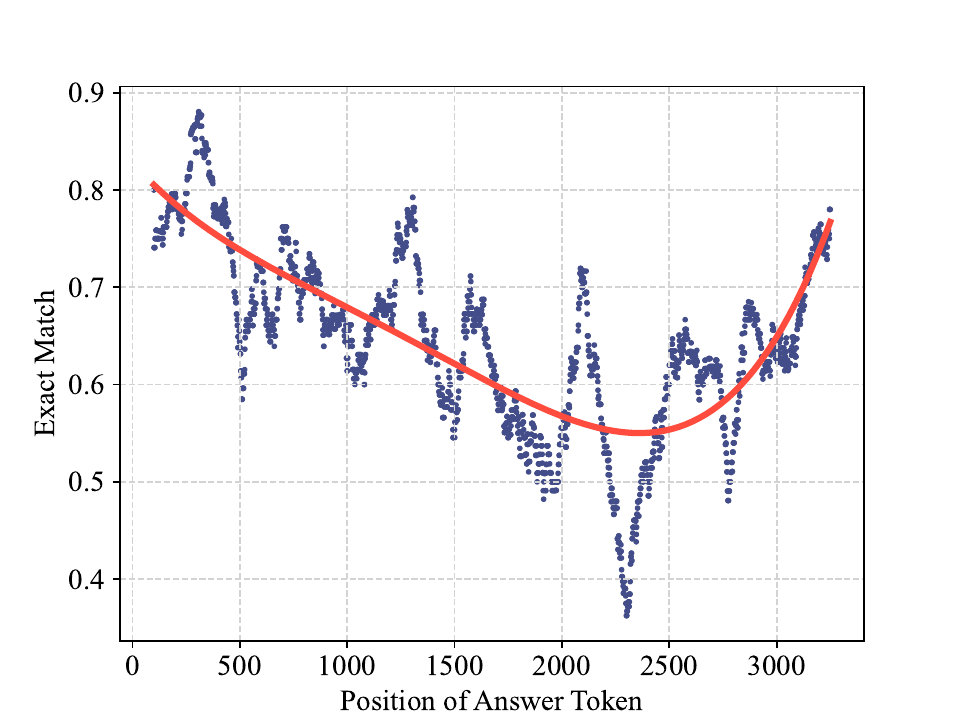}
    }
    
    \subcaptionbox{CodeLlama\label{fig:llama_dot2}}{
        \includegraphics[width=0.4\textwidth]{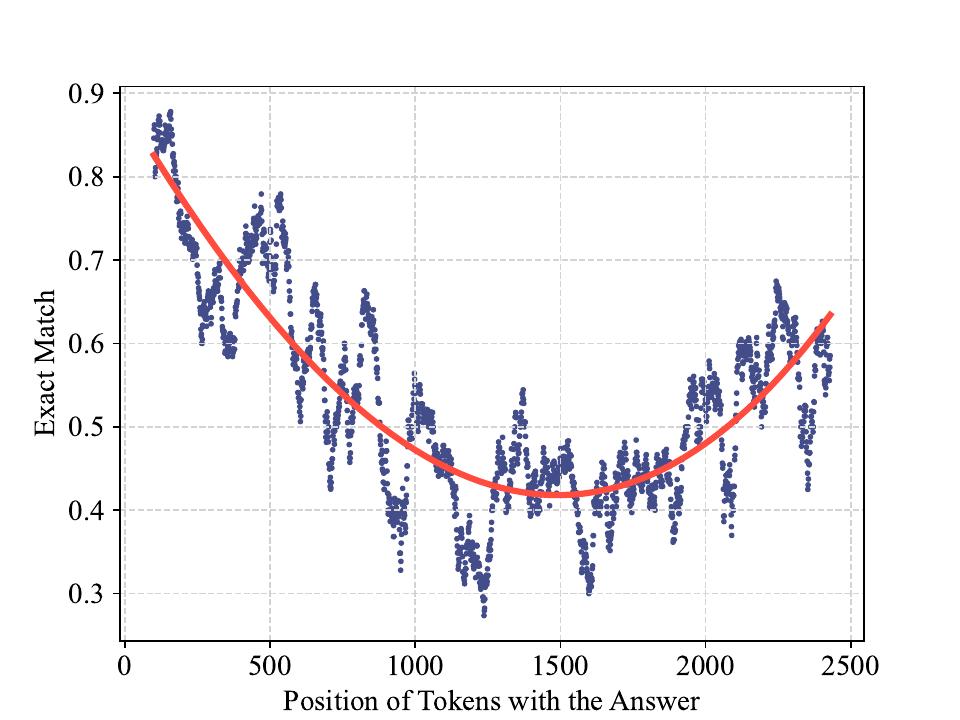}
    }

    \caption{The relationship between LLMs performance and the position of the answer token. 
    }
    
    \label{fig:lost_in_the_middle}
\end{figure}

\begin{figure*}[tb]
    \centering
	\includegraphics[width=0.7\textwidth]{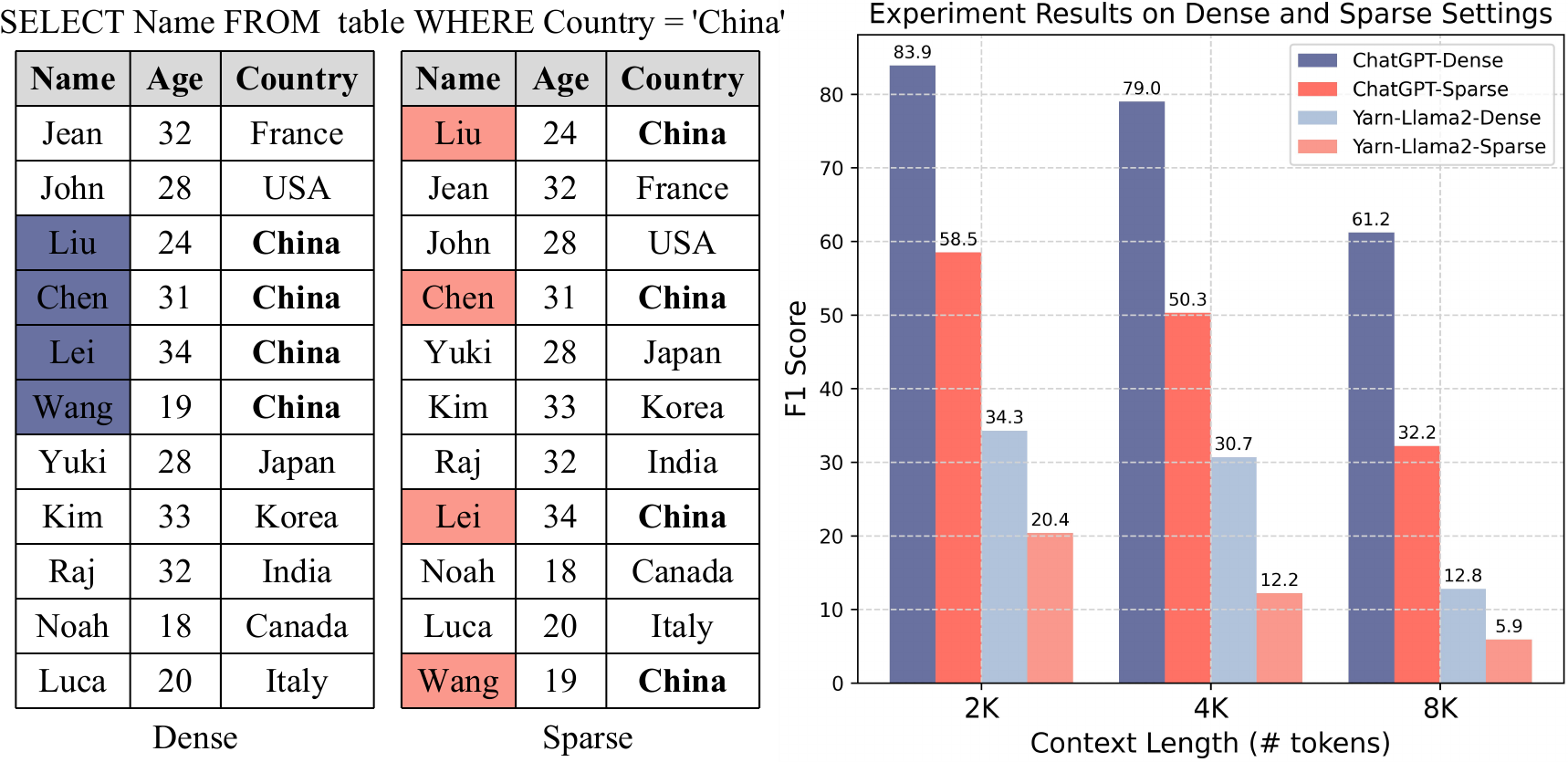}
	\caption{Experiment results of ChatGPT and Yarn-Llama 2 on \textit{Dense} and \textit{Sparse} Settings. \textit{Dense} means that the answer cells (i.e., \texttt{\small Liu}, \texttt{\small Chen}, \texttt{\small Lei}, \texttt{\small Wang}) lie in adjacent rows, and \textit{Sparse} means that the answer cells are separated. The model performs better on local queries which only involves adjacent cells.}
	\label{fig:neighbor}
\end{figure*}

\section{Systematic Suite: Controllable Analysis}
\label{sec:systematic}

\eval provides a comprehensive framework that empowers developers to synthesize diverse evaluation examples for systematically assessing LLMs from multiple perspectives.
In this section, inspired by the work of lost in the middle~\citep{liu2023lost}, we first analyze the impact of answer position on performance (Section~\ref{sec:answer_position}).
Then we evaluate LLMs from different viewpoints, and we have conducted some initial explorations on the reasoning types analysis (Section~\ref{app:reasoning_type}).
Last, we provide some insights by analyzing LLMs on three selected SQL templates (Section~\ref{app:template_control_analysis}).
These experiments reveal counter-intuitive performance trends and new discoveries that may inspire further research and extension of the work.

\subsection{Answer Position Analysis}
\label{sec:answer_position}
We investigate the influence of the answer's position on the performance of LLMs, which is generally considered important. 
Unlike standard NLP benchmarks where it is difficult to control the position of the answer, \eval allows for fine-grained control of answer position at the token level.
To mitigate the influence of long contexts, we only analyzed answers that fell within a limited context window (i.e., less than 4K tokens).

Echoing the findings of \citet{liu2023lost}, ``lost in the middle'', our results in Figure~\ref{fig:lost_in_the_middle} demonstrate that both ChatGPT and CodeLlama achieve higher performance when the answer is located at the beginning or end of the context, compared to when it appears in the middle.
In addition, we found a periodic fluctuation trend in the performance of both models as the position of the answer shifts within the context.
For example, the performance of ChatGPT increases from 0 to around 200, then starts to decrease from around 200 to 500. 
This wave-like pattern in performance appears to correlate with the position embedding approach used by LLMs.

In contrast to previous studies that used long-context question answering tasks~\citep{liu2023lost,bai2023longbench} for analysis and are thus limited to controlling answer positions at the paragraph level, \eval provides a more precise approach by focusing on token level. This key difference enables \eval to offer fine-grained control and promote the exploration of relevant phenomena.

\subsection{Answer Distribution Analysis}
Given the limitation of existing LLMs on long-context tasks, we are curious about the bottleneck of them.
By using \eval, we can systematically investigate the long-context modeling capabilities of LLMs by controlling the distribution of answers in the evaluation suite.
Specifically, we use the \textbf{Easy} setting and fix the number of answers to four cells (i.e., the result of the SQL execution is always spanning four cells).
As illustrated in Figure~\ref{fig:neighbor}, we introduce two distribution patterns, \textit{Dense} and \textit{Sparse}~\footnote{Examples of these two patterns can be found in Appendix~\ref{app:neighbor_disperse}.} to probe the limitations of current LLMs.
The dense mode only requires the model to understand the local context, whereas the sparse mode requires the model to have a broader, global understanding of the context across multiple blocks.
The sparse mode intuitively poses more challenges and demands more complex reasoning across a broader scope of the provided context.
We conduct experiments on ChatGPT and Yarn-llama2-13B~\citep{peng2023yarn}.
The experimental results indicate that both models perform significantly better in dense mode compared to sparse mode, as shown in Figure~\ref{fig:neighbor}.
This indicates that LLMs struggle to retrieve information over long sequences, even though their pre-training included lengthy contexts.
This may be caused by the fact that the training data does not contain sufficient examples of long-distance dependencies for the model to learn effectively.
Furthermore, the steep drop in performance from 4K to 8K tokens for both ChatGPT and Yarn-Llama2 in dense mode indicates that current length extension techniques may not be as effective as hoped.
In summary, we believe that \eval provides a valuable framework for evaluating long-context language models, as it allows testing models on dialogues of arbitrary length. This establishes a solid foundation for advancing research on large language models that can leverage long-term context.

\subsection{Reasoning Type Analysis}
\label{app:reasoning_type}

\definecolor{orange}{RGB}{255, 165, 0}
\definecolor{bittersweet}{RGB}{255, 102, 102}
\definecolor{green}{RGB}{0, 128, 0}
\definecolor{cyan}{RGB}{0, 255, 255}
\definecolor{plum}{RGB}{221, 160, 221}
\definecolor{blueviolet}{RGB}{138, 43, 226}

\begin{table*}[htbp]
\centering
\tiny
\begin{tabular}{cp{3.8cm}cccc}
\toprule
\textbf{Operator} &  \textbf{Example SQL} & \textbf{ChatGPT} & \textbf{Claude} & \textbf{Mistral} & \textbf{CodeLlama} \\
\midrule
{\textcolor{orange}{\textbf{Filter}}} & SELECT lyonnais FROM table WHERE farmer = 'mijl' AND lashing \textgreater 288  &  79.6 & 79.2 & 64.8 & 72.8 \\
\midrule
{\textcolor{green}{\textbf{Arithmetic}}} & SELECT synset + refuge FROM table WHERE blender = 'owxdbzjg' & 67.2 & 59.4 & 5.4 & 10.6\\
\midrule
{\textcolor{plum}{\textbf{Comparative}}} & SELECT upsetter \textless{} jollity FROM table WHERE kelp = 150 & 45.2 & 46.4 & 44.8 & 46.6 \\
\midrule
{\textcolor{bittersweet}{\textbf{Aggregate}}} & SELECT MIN(skeptic) FROM table & 38.4 & 39.4 & 28.4 & 33.8 \\
\midrule
{\textcolor{blueviolet}{\textbf{Group}}} & SELECT lats FROM table GROUP BY shastan HAVING sum ( logbook ) = 56 & 38.1 & 28.2 & 31.0 & 37.8 \\
\midrule
{\textcolor{cyan}{\textbf{Superlative}}} & SELECT severity FROM table ORDER BY bierce DESC Limit 1 & 24.8 & 41.4 & 19.2 & 28.3\\
\bottomrule
\end{tabular}
\caption{Reasoning types experiments examples of different LLMs.}
\label{tab:reasoning}
\end{table*}

\eval enables the creation of multiple templates to generate different SQL statements, with each statement representing a distinct reasoning type.
We selected six common reasoning types to investigate the reasoning capabilities of LLMs and examined four different LLMs: ChatGPT, Claude, Mistral-7B, and CodeLlama-34B.
Following~\citet{liu2021tapex}, the six reasoning types~\footnote{Detailed templates for each type can be found in Appendix~\ref{app:template}.} we considered are \texttt{Filter}, \texttt{Aggregate}, \texttt{Arithmetic}, \texttt{Superlative}, \texttt{Comparative}, and \texttt{Group}. The example SQL and the experimental results of different LLMs are presented in Table~\ref{tab:reasoning}.
The expressive power of SQL queries enables \eval to be used for evaluating diverse scenarios such as numerical reasoning, multi-hop reasoning, complex code understanding, and multi-turn interaction with intermediate execution results.

\section{Related Work}

Evaluating large language models (LLMs) has garnered significant interest in the NLP community~\citep{chang2023survey}. This allows us to gain a deeper understanding of the specific capabilities and limitations of LLMs while guiding further research.
Researchers proposed MMLU~\citep{hendrycks2020measuring} to measure the knowledge acquired by a language model during pre-training. In recent years, with the development of LLMs, a series of general evaluation benchmarks have emerged. For instance, BBH~\citep{suzgun2022challenging} and AGIEval~\citep{zhong2023agieval} assess the reasoning ablitities. GSM8K~\citep{cobbe2021training} evalutes the math reasoning, HumanEval~\citep{chen2021evaluating} and MBPP~\citep{austin2021program} measure code capalities. Our work aims to provide an evaluation suite for measuring reasoning ability.

Many previous works on long-text modeling rely on the perplexity ~\citep{sun2021long, peng2023yarn} or performance on simple artificial tasks~\citep{li-roth-2002-learning, berant2013semantic, mohtashami2023landmark}. Concurrently, ZeroSCROLLS~\citep{shaham2023zeroscrolls}, L-Eval~\citep{an2023eval} and LongBench~\citep{bai2023longbench} are proposed as evaluation benchmarks for long-text modeling. However, these benchmarks are built from existing public datasets and have fixed evaluation types. In contrast, \eval can effectively assess comprehension of infinitely long-context. Furthermore, \eval allows customization of settings to generate evaluation data that meets specific needs, enabling effective evaluation of model deficiencies and discovery of new insights into LLMs.

\section{Conclusion}

In this paper, we have introduced \eval, a novel synthetic evaluation suite for LLMs using SQL execution. \eval represents a scalable and systematic approach to evaluate LLMs on a dynamic task.
Our experiments demonstrate strong correlation between \eval and traditional evaluation benchmarks.
The key innovations of \eval are its flexibility, allowing unlimited context length and unlimited evaluation examples, and its fine-grained, systematic nature which enables detailed analysis of model capabilities and flaws.

Most importantly, for long-context evaluation, \eval can generate evaluation data of infinite length. This type of task is not only challenging but also rooted in real-world tasks. Considering the rapid development of LLMs, even as LLM lengths extend significantly, \eval can serve as a valuable benchmark for LLM development and contribute to the community.

\section*{Limitations}
Besides the features described in this paper, it currently supports complex multi-turn SQL execution task and multi-turn instruction task. Moreover, it also supports multilingual testing, especially for reasoning data generation of low-resource languages, which has not been widely studied by the academic community.
However, this paper has not yet conducted a systematic analysis of these complex new features.

In addition, due to the complex and diverse syntax of SQL, the syntax that \eval can generate is still relatively limited, which is also what we need to do in our future work. Moreover, there is currently no toolkit that can randomly generate a large number of complex SQLs, which is also a significance of our work.

Due to space limitations, many valuable experimental results are shown in Appendix~\ref{app:control}. We analyzed in detail the impact of various types of influencing factors on the results and have drawn other valuable conclusions.

Exploring the treasure contained in synthetic data is our goal for the future, and we believe that this work can bring inspiration to this field.

\section*{Acknowledgements}
This work was supported by the National Key R\&D Program of China (No.2022ZD0160503) and the National Natural Science Foundation of China (No.62376270), Youth Innovation Promotion Association CAS, and OPPO Research Fund.

\bibliography{custom}

\newpage
\appendix

\section{Evaluation Experiments Results}
\label{app:detail_analysis}

\subsection{Other Synthetic Task}
\label{app:other_synthetic}
\begin{figure}[h]
    \centering
 \includegraphics[width=0.4\textwidth]{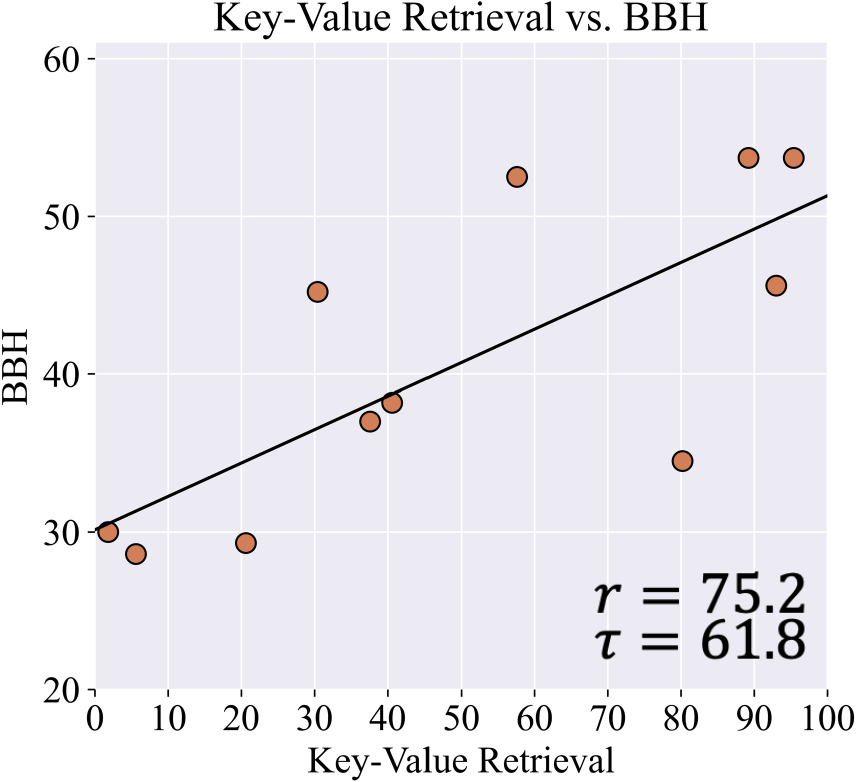}
 \caption{Performance analysis of key-value retrieval task and BBH.}
 \label{fig:pkv_bbh}
\end{figure}

\eval is a synthetic task that possesses a certain level of difficulty and robustness, which allows for a good assessment of an LLM's overall capability compared to previous works. We choose key-value retrieval task~\citep{liu2023lost}, given a key, the goal is to return the associated value. We test several LLMs on this task, and the experiments results are shown in Figure~\ref{fig:pkv_bbh}. It demonstrates that key-value retrieval task is a simple task which has low correlation with real LLMs reasoning benchmark. \eval, as a complex and robust benchmark, can provide reference for future synthetic data.

\subsection{Overall Performance}
\label{app:overall performance}
The detail performance are shown in Table~\ref{tab:mail_results}.

\begin{table*}[t]
\centering
\small
\begin{tabular}{clccccc}
\toprule
\multicolumn{1}{l}{}         &       & \multicolumn{2}{c}{\textbf{Synthetic Task}} & \multicolumn{2}{c}{ \textbf{Realistic Benchmark}} &  \\
\cmidrule{3-4} \cmidrule{5-6}
\multicolumn{1}{l}{}         &        & \eval-Easy             & \eval-General                        & WTQ            &   Reasoning Task                    &                            \\ \midrule
\multirow{21}{*}{LLM} & GPT-4           & \textbf{99.4}    & \textbf{63.1}                       & 70.8       & 86.7                                 \\
                             & ChatGPT    & 97.0             & 47.2                         & 62.0          & 70.1                                 \\
                             & Claude-1    & 98.2             & 44.3                             & 63.4          & 67.3                                                                            \\
                             & Llama-2-70B     & 94.2             & 41.3                         & 55.9         & 64.9                                                                            \\
                             & Mistral-7B      & 87.4             & 34.3                         & 55.7          & 53.7                                                                        \\
                             & Llama2-13B      & 75.0             & 30.9                         & 49.2          & 45.6                                                                            \\
                             & InternLM-20B    & 78.0             & 32.3                         & 49.4          & 52.5                                                                            \\
                             & Qwen-14B        & 71.8             & 25.8                         & 46.7          & 53.7                                                                            \\
                             & Llama-2-7B      & 54.2             & 23.8                         & 40.6          & 38.2                                                                            \\
                             & Qwen-7B         & 56.4             & 19.4                         & 41.2          & 45.2                                                                            \\
                             & Xgen-7B         & 55.2             & 24.6                         & 36.3          & 34.5                                                                               \\
                             & Internlm-7B     & 41.6             & 18.5                         & 27.5          & 37.0                                                                            \\
                             & Phi-1\_5        & 27.6             & 16.1                         & 22.1          & 30.0                                                                         \\
                             & Stablelm-7B     & 6.0              & 4.2                          & 14.7          & 24.3                                                                         \\
                             & Stablelm-3B     & 4.2              & 2.9                          & 11.2          & 21.0                                                                         \\
                             & Pythia-12B      & 31.4             & 17.3                         & 24.5          & 29.3                                                                         \\
                             & Pythia-6.9B     & 25.2             & 16.0                         & 22.6          & 28.6                                                                               \\
                             & Pythia-2.8B     & 26.4             & 14.6                         & 21.7          & 28.8                                                                              \\
                             & Pythia-1B       & 8.4              & 7.1                          & 16.2          & 25.6                                                                               \\ \midrule
\multirow{10}{*}{Code LLM}   & CodeLlama-34B   & 91.4             & 41.0                         & 53.9                      & 36.4                                                                   \\
                             & CodeLlama-13B   & 90.0             & 35.7                         & 49.9                      & 30.6                                                                   \\
                             & CodeLlama-7B    & 75.2             & 34.2                         & 44.9                      & 26.3                                                                   \\
                             & StarCoder-15B   & 87.2             & 34.4                         & 39.2                      & 30.4                                                                   \\
                             & StarCoder-7B    & 88.4             & 32.4                         & 33.3                      & 28.3                                                                   \\
                             & StarCoder-3B    & 79.0             & 28.0                         & 27.5                      & 21.5                                                                   \\
                             & StarCoder-1B    & 37.4             & 15.4                         & 21.1                      & 15.2                                                                   \\
                             & CodeGen-15B     & 36.8             & 18.2                         & 25.0                      & 18.3                                                                   \\
                             & CodeGen-6B      & 25.0             & 16.9                         & 17.8                      & 18.2                                                                   \\
                             & CodeGen-2B      & 31.4             & 16.6                         & 20.8                      & 14.5                                                                  \\ \bottomrule
\end{tabular}
\caption{SQL Execution Task Performance on different LLMs.}
\label{tab:mail_results}
\end{table*}

\subsection{Reliability Experiments}
\label{app:alignment}
\begin{figure*}[tb]
    \centering
    \subcaptionbox{Correlation between QA task and SQL execution Task on WikiTableQuestions. \label{fig:sql_qa}}{
        \includegraphics[width=0.3\textwidth]{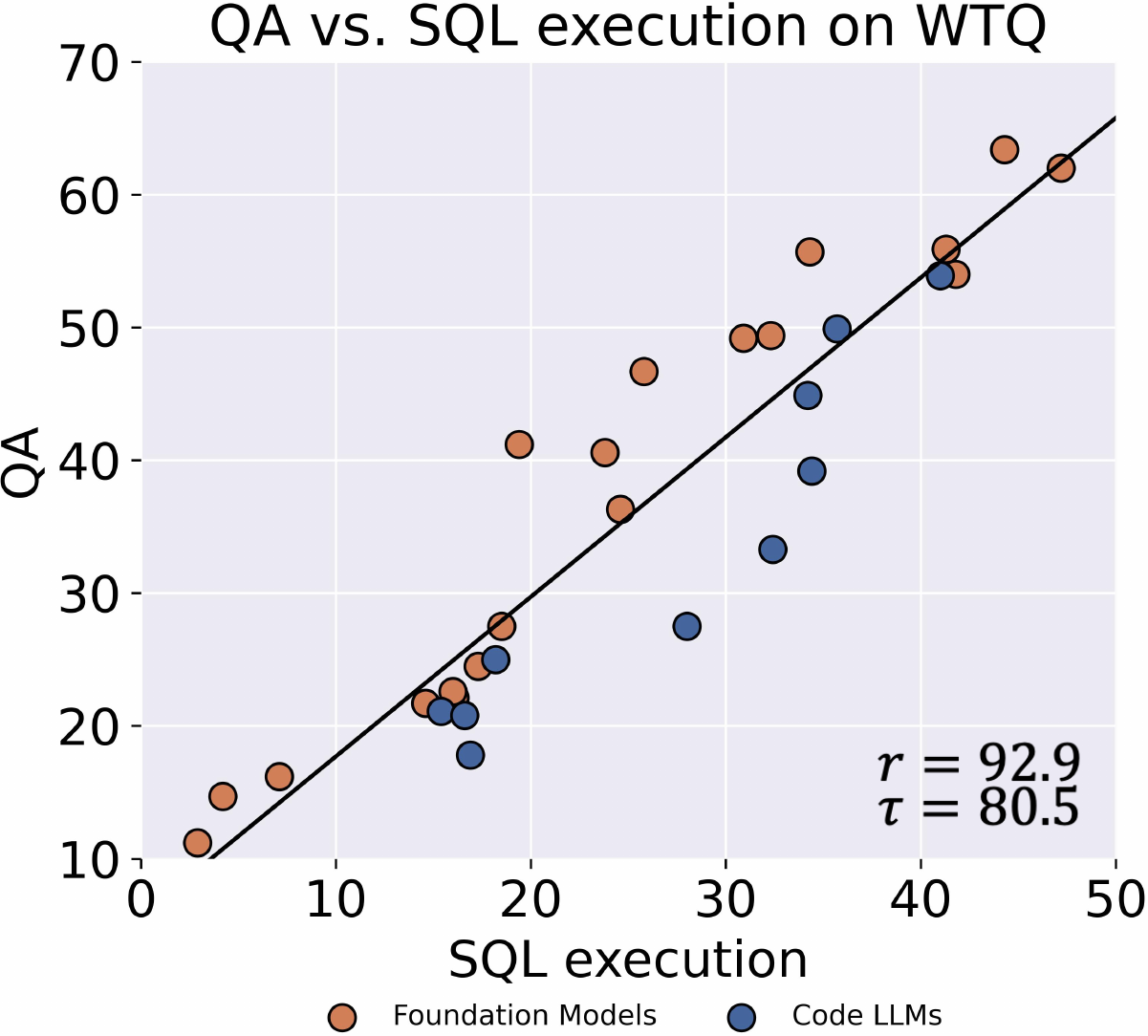}
    }
    \hfill
    \subcaptionbox{Correlation between General and Easy Settings.\label{fig:general_easy}}{
        \includegraphics[width=0.31\textwidth]{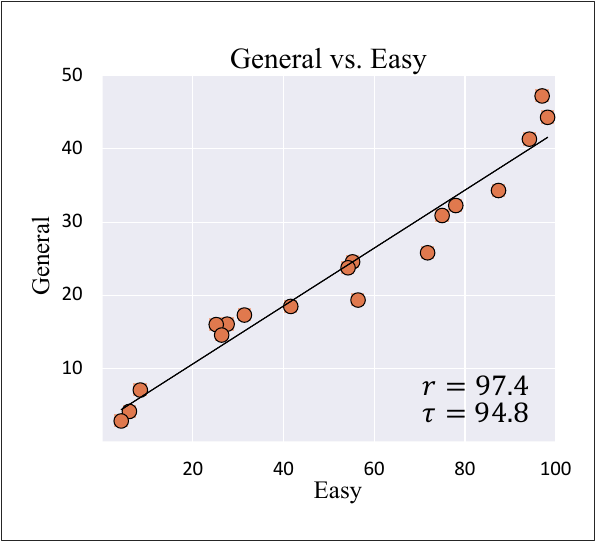}
    }
    \hfill
    \subcaptionbox{Correlation between Synthetic and Real Table SQL execution task\label{fig:wtqsql_general}}{
        \includegraphics[width=0.32\textwidth]{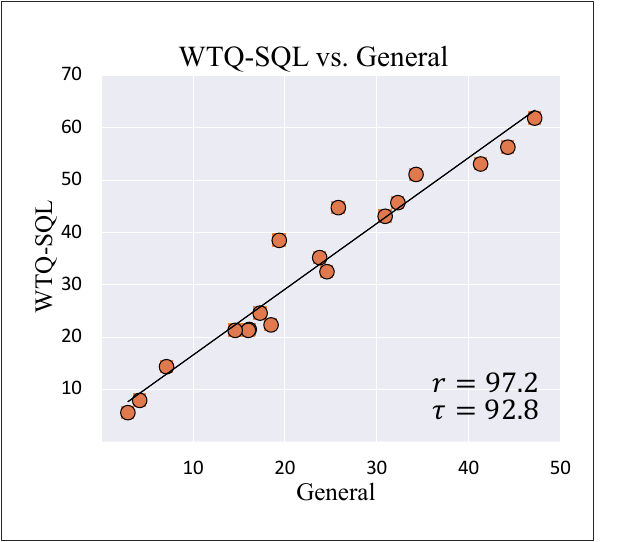}
    }

\caption{Experimental results of the correlation experiments.}
\end{figure*}

\paragraph{Symbolic Tasks vs. Natural Language Tasks.}
Another point to prove is that symbolic tasks are consistent with their natural language counterparts. SQL execution is a suitable task because SQL can be intertranslated with an natural question. As can be seen from the ``WTQ'' column of the Table~\ref{tab:mail_results} and Figure~\ref{fig:sql_qa}, LLM's ability to execute SQL is consistent with its table question answering ability. 

\paragraph{Synthetic data vs. Real data.} We want to verify if the synthesized SQL is simpler. The tables ``SQL-general'' and ``WTQ-SQL'' show the difference in performance between the model on synthetic and real data. We keep the average length of the tables similar, and the experimental results show that the synthetic SQL is more complex than the real SQL. And Figure~\ref{fig:wtqsql_general} shows that, the performance of LLMs on real tables and synthetic tables is very relevant.

\paragraph{Different \eval Settings.} As shown in Figure~\ref{fig:general_easy}, even if the data settings are very different, LLMs are guaranteed a consistent performance ranking on \eval.

\subsection{Other SQL Prompting Styles}

\paragraph{SQL execution task with Chain-of-Thought prompting.}
SQL is a complex multi-step reasoning task. To verify whether it is a reliable reasoning task, \eval generates multi-step execution instructions for SQL. ChatGPT's performance~(markdown) improves from \textbf{38.0} to \textbf{48.5} when using chain-of-thougnt prompts. The chain-of-thought examples are shown in below. The examples of chain-of-thought prompting are shown in Appendix~\ref{app:chain_of_thought}.

\paragraph{SQL multi-step instruction experiments.} SQL multi-step instruction is an auxiliary task. We generate two new datasets using different settings than \textbf{Easy} and \textbf{General}, named \textbf{Data1} and \textbf{Data2}. Experiments results are shown in Table~\ref{tab:multistep}.

\subsection{Long-Context Experiments}
\label{app:long_context}

Context windows limit the long-context capabilities of LLMs. Previous researchers have proposed many ways to extend the length of context windows, often to 64K, 128K and so on. Existing benchmarks~\citep{bai2023longbench, an2023eval} collect data from existing NLP communities (which causes data leakage), and more importantly because collecting large amounts of data is difficult. \eval, on the other hand, is easy to collect data with variety and complexity. Existing benchmarks also can't effectively evaluate very long texts, but \eval can evaluate arbitrary lengths. 

YaRN~\citep{peng2023yarn} extend LLaMA2 context windows to 128K, however, they only evaluated the model's perplexity, which we believe is not a true reflection of its long-context understanding capability. So we use \eval to generate table data of different lengths and keep all parameters same to evaluate the performance of yarn-LLaMA2, and the experimental results are shown in Table~\ref{tab:long_context}. It shows that, yarn-llama2 has a noticeable dip in performance on 20K-80K, which is good for a small number of tasks as well. But compared to ChatGPT (which we can only test 16K length tables), there's a noticeable gap.
\begin{table*}[t]
\centering
\begin{tabular}{lccccccccc}
\toprule
\multicolumn{1}{c}{\multirow{2}{*}{Model}} & \multirow{2}{*}{Max-Ctx} & \multicolumn{8}{c}{SQL Execution}              \\
\multicolumn{1}{c}{}                       &             & 2K         & 4K   & 8K   & 16K  & 20K  & 40K  & 60K  & 80K  \\ \midrule
ChatGPT                              & 16k         & 96.8     & 95.2 & 80.3 & 68.7 & -    & -    & -    & -    \\
Claude-1.3-100K                            & 128k        & 97.2     & 96.8 & 91.8 & 85.2 & -    & -    & -    & -    \\
Yarn-LLaMA2-13B                            & 128k        & 76.3     & 57.0 & 40.6 & 25.1 & 20.6 & 17.6 & 17.0 & 12.0 \\
XGen-7B                                    & 8k          & 51.6     & 41.8 & 25.4 & -    & -    & -    & -    & -    \\
LongChat-13B                               & 16k         & 48.6     & 39.0 & 26.3 & 19.5 & -    & -    & -    & -    \\ 
LongLlaMA-7B                               & 256k        & 82.4 & 62.8 & 24.4 & -    & -    & -    & -  & -    \\ 
RWKV-Raven-14B                             & 128k        & 10.5          & 7.4 & 6.2 & - & -    & -    & -    & -    \\ \bottomrule
\end{tabular}
\caption{Long-Context experiments on \eval.}
\label{tab:long_context}
\end{table*}

\begin{table*}[t]
\centering
\small
\begin{tabular}{@{}lcccccccc@{}}
\toprule
\multirow{3}{*}{\textbf{Model}} & \multicolumn{4}{c}{\textbf{SQL Execution}}                                           & \multicolumn{4}{c}{\textbf{SQL Multi-Step Instruction}}                                            \\
                                & \multicolumn{2}{c}{\textit{Zero-Shot}} & \multicolumn{2}{c}{\textit{Few-Shot}} & \multicolumn{2}{c}{\textit{Zero-Shot}} & \multicolumn{2}{c}{\textit{Few-Shot}} \\
                                & \textbf{Data1}              & \textbf{Data2}             & \textbf{Data1}              & \textbf{Data2}              & \textbf{Data1}              & \textbf{Data2}             & \textbf{Data1}              & \textbf{Data2}              \\        
\midrule
ChatGPT       & 96.4 & 47.0 & 97.0 & 49.0 & 97.9 & 30.0 & 98.8 & 34.8 \\
Codellama-13B      & 71.2 & 34.3 & 90.0 & 39.8 & 63.9 & 12.1 & 88.0 & 22.8 \\
StarCoder-15B      & 52.3 & 24.7 & 85.8 & 37.6 & 44.4 & 14.4 & 84.2 & 19.2 \\
InternLM-20B     & 60.4 & 22.7 & 78.0 & 35.0 & 58.8 & 14.9 & 76.6 & 28.1 \\
InternLM-20B-Chat     & 71.2 & 31.3 & 78.0 & 34.2 & 67.6 & 21.9 & 74.4 & 25.4 \\
LLaMA2-13B      & 68.1 & 23.2 & 75.0 & 32.3 & 50.5 & 5.4 & 74.6 & 18.2 \\
LLaMA2-13B-Chat      & 51.6 & 16.4 & 71.5 & 28.3 & 9.4 & 1.0 & 64.2 & 21.1 \\
Vicuna-13B      & 57.6 & 26.8 & 81.6 & 35.4 & 48.9 &11.5 & 78.8 & 24.2 \\
\bottomrule
\end{tabular}
\caption{SQL Multi-Step Task performance on different LLMs.}
\label{tab:multistep}
\end{table*}

\section{Controllable Analysis Results}
\label{app:control}
\subsection{Answer Position Analysis}
\begin{figure*}[tb]
    \centering
    \subcaptionbox{\label{fig:gpt_dot}}{
        \includegraphics[width=0.45\textwidth]{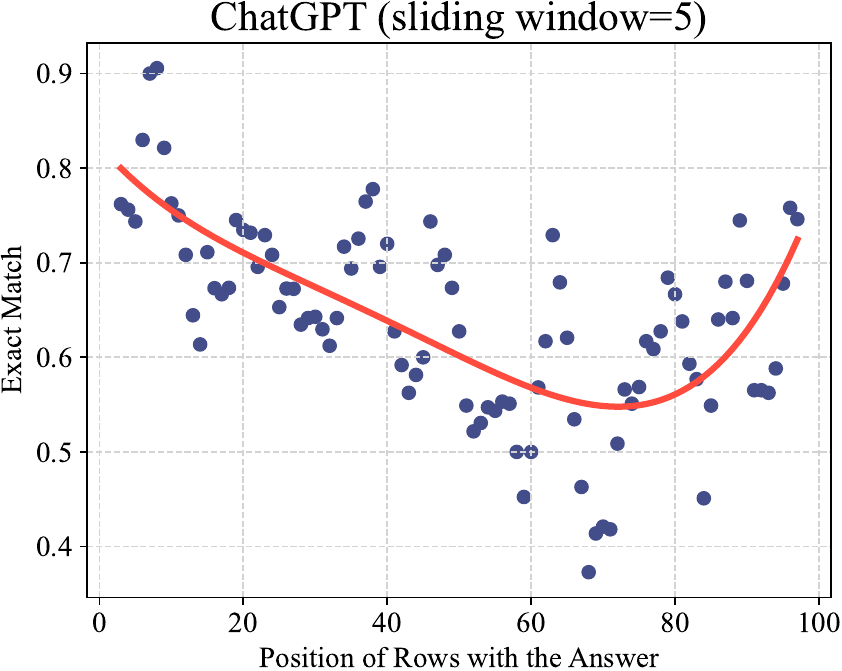}
    }
    \hfill
    \subcaptionbox{\label{fig:llama_dot}}{
        \includegraphics[width=0.45\textwidth]{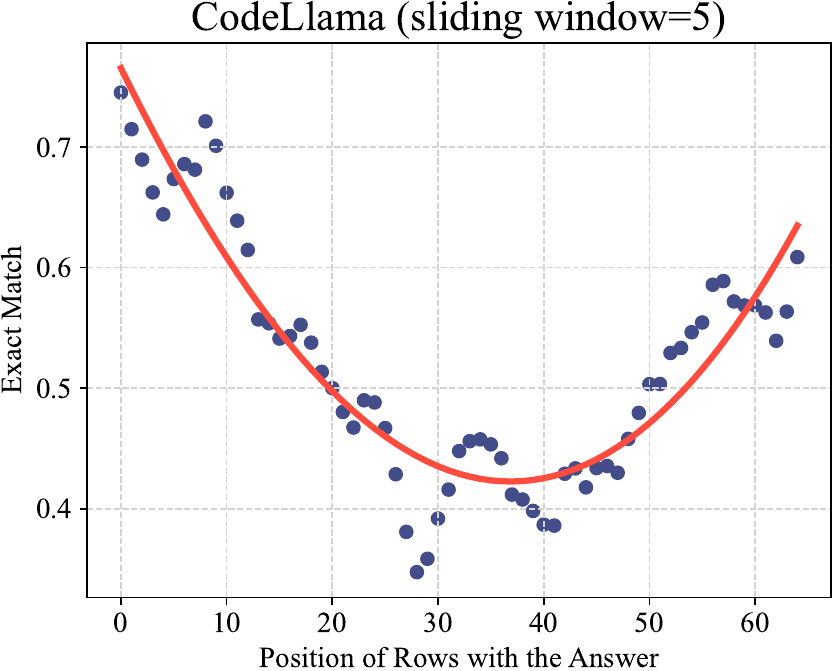}
    }

    \subcaptionbox{\label{fig:gpt_group}}{
        \includegraphics[width=0.45\textwidth]{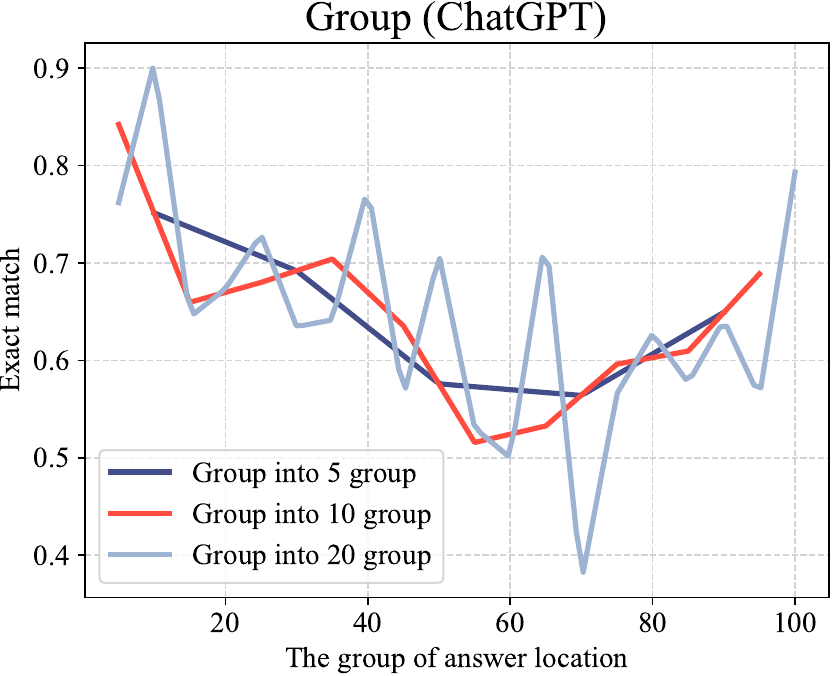}
    }
    \hfill
    \subcaptionbox{\label{fig:llama_group}}{
        \includegraphics[width=0.45\textwidth]{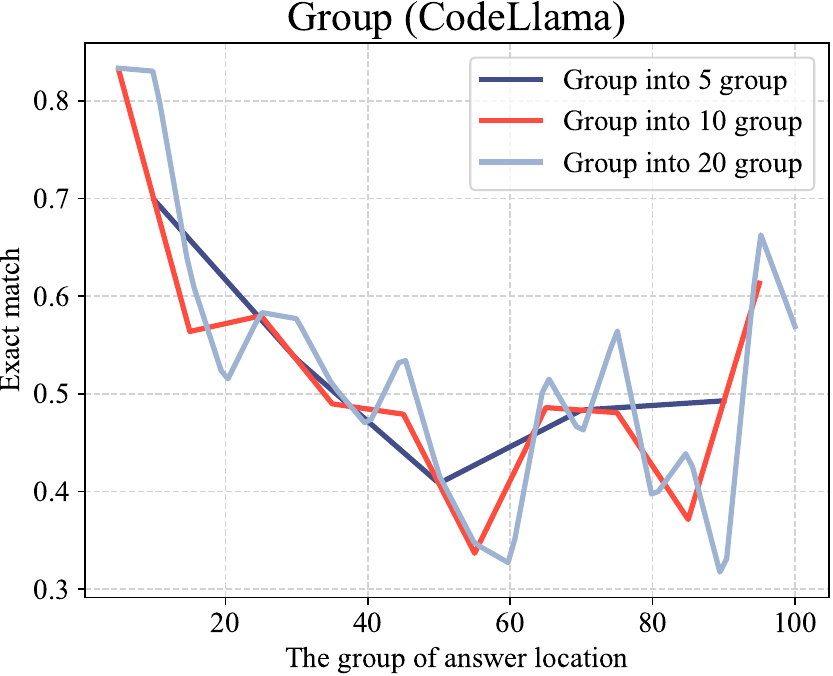}
    }
\caption{Effect of answer position on model performance. We use two methods to visualize the results. (1) Sliding windows~(Figure~\ref{fig:gpt_dot},\ref{fig:llama_dot}). We select a window size of 5 and smooth the data to make a dot plot and a trend line. (2) Grouping calculations~(Figure~\ref{fig:gpt_group},\ref{fig:llama_group}). We group neighboring rows with granularities of 5, 10, and 20. For instance, with a granularity of 20, we group rows with answers located in the ranges 1-20, 21-40, 41-60, 61-80, and 81-100, resulting in five groups, and compute the average scores.}
\end{figure*}

In addition to the figures in the main text, we also conduct experiments with row level. We use two methods to visualize the results. (1) Sliding windows (Figure~\ref{fig:gpt_dot},\ref{fig:llama_dot}). We choose windows=5 and smooth the data to make a dot plot and a trend line. (2) Grouping calculations~(Figure~\ref{fig:gpt_group},\ref{fig:llama_group}). Group neighboring rows together with the granularity of 5, 10, and 20. For example, if granularity is 20, then we group the rows with answers located in 1-20, 20-40, 40-60, 60-80, and 80-100, for a total of five groups, and calculate the average scores.

\subsection{Template Controlled Analysis}
\label{app:template_control_analysis}

Each data template in \eval includes corresponding reasoning types, and thus it provides fine-grained control over the evaluation examples.
To stimulate new insights and uncover counter-intuitive performance phenomena of LLMs, we present several controlled analysis examples using simple templates as a starting point.

\textbf{Template1: SELECT \text{[}text\_col1\text{]} FROM table WHERE (\text{[}text\_col2\text{]} = \text{[}text2\text{]})}

\begin{figure*}[t]
  \centering
\begin{minipage}{0.32\textwidth}
    \centering
    \includegraphics[width=\textwidth]{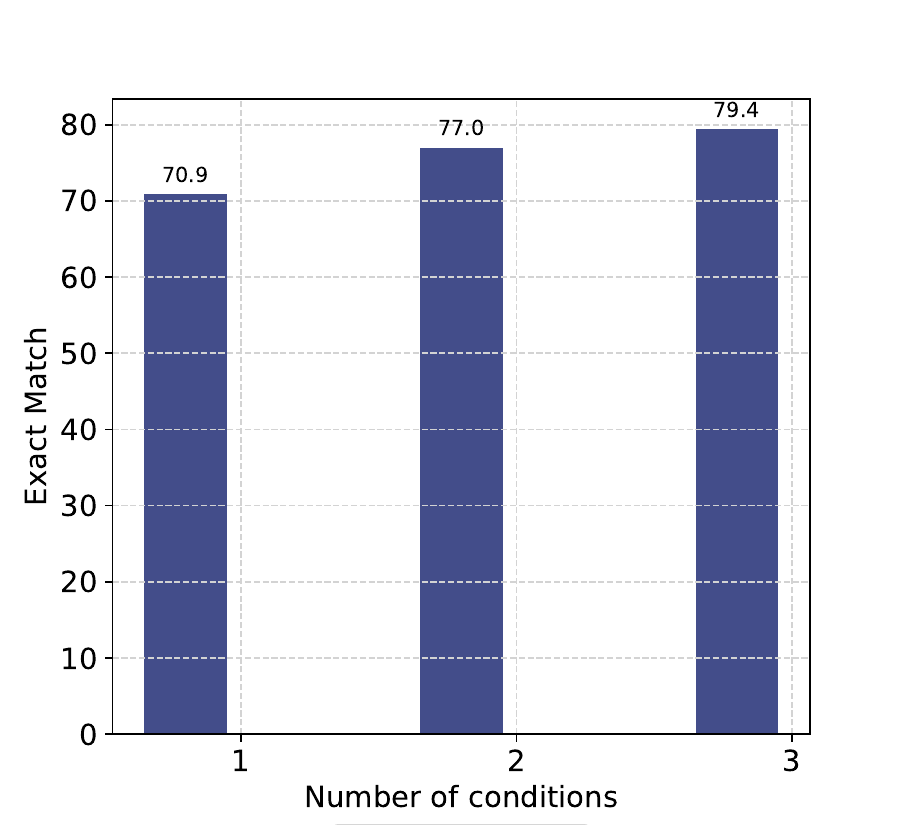}
    \caption{Trend of ChatGPT performance with where condition number using Template2.}
    \label{fig:where_num}
  \end{minipage}
\hfill
  \begin{minipage}{0.32\textwidth}
    \centering
    \includegraphics[width=\textwidth]{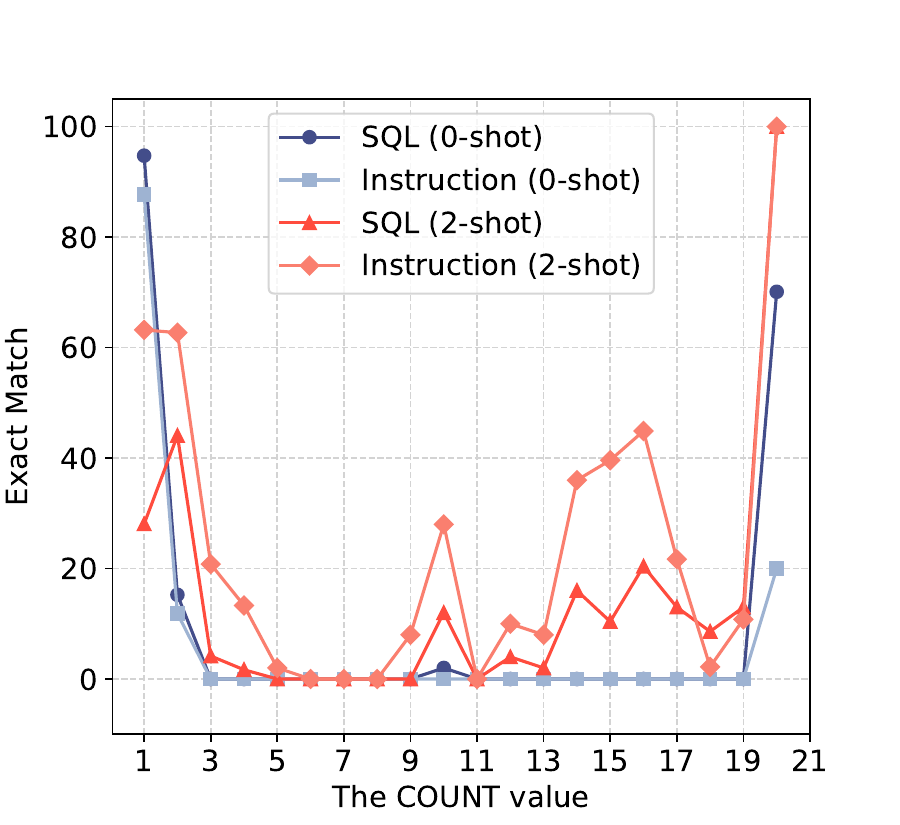}
    \caption{Trend of ChatGPT performance with the COUNT value in Template3. Only when the COUNT value is the largest or smallest, the model have good performance.}
    \label{fig:count}
  \end{minipage}
\hfill
  \begin{minipage}{0.32\textwidth}
    \centering
    \includegraphics[width=\textwidth]{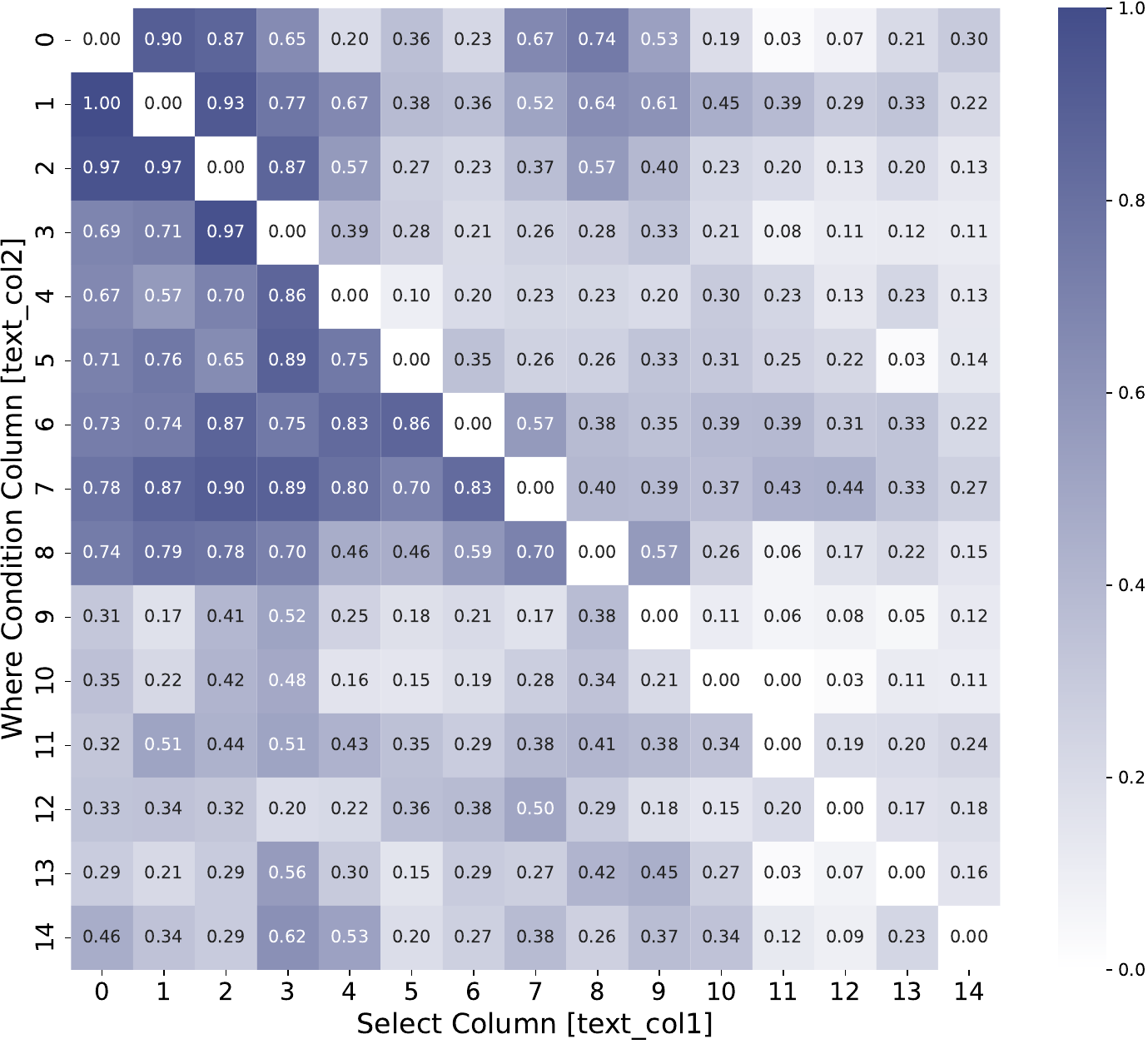}
    \caption{ChatGPT performance with different locations of  [text\_col1] and [text\_col2]. The performance improves when the example has the location of [text\_col1] before [text\_col2].}
    \label{fig:heatmap}
  \end{minipage}
  
\end{figure*}

We first explore the relationship between the model performance and the locations of [text\_col1] and [text\_col2].
To begin with, we generated a set of $10\times15$ tables, each comprising $15$ distinct columns. 
We created 400 unique combinations by pairing each value in text\_col1 with each value in text\_col2. For each of the 400 pairs, we generated 40 evaluation examples, resulting in a total of 16,000 evaluation examples.
After SQL execution experiments, we calculated the scores of each pair and constructed a heatmap, which is illustrated in Figure~\ref{fig:heatmap}. The heatmap indicates that the performance is overall better when [text\_col1] is the previous column. And the model performance is also better when the [text\_col1] column is before [text\_col2] column. 
It indicates that the model tends to focus on the beginning of a specific paragraph. Moreover, in multi-hop reasoning, LLMs excel at hopping to the context preceding a intermediate hop, but struggles when it comes to searching backward.

\textbf{Template2: SELECT \text{[}text\_col1\text{]} FROM table WHERE (\text{[}text\_col2\text{]} = \text{[}text2\text{]}) $\times$ N}

We then investigate the impact of the number of WHERE conditions on LLM performance. Intuitively, more conditions should make it harder for LLM to execute SQL since the instruction becomes more complex.
However, the experimental results contradict this intuition, as shown in blue in Figure~\ref{fig:where_num}. We speculate that this counter-intuitive result stems from how LLMs actually reason: by looking up string co-occurrences rather than logically considering all conditions.

\textbf{Template3: SELECT COUNT(\text{[}text\_col\text{]}) FROM table WHERE \text{[}text\_col\text{]} = \text{[}text\text{]} }.

We analyze the counting ability of LLMs, which is an important numerical reasoning capability.
To avoid potential symbolic effects of SQLs, we also use the instruction style~(Section~\ref{sec:method}) to prompt the model~(e.g. Please count the number of ``\text{[}text\_col\text{]} is \text{[}text\text{]}''). As shown in Figure~\ref{fig:count}, whether it is zero-shot or few-shot, SQL style or instruction style, the performance of LLMs is best when the COUNT value is the smallest or the largest. When the COUNT value is in the middle, the performance of the model is almost zero. 

In the future, developers can employ the \eval suite to analyze the performance of LLMs with various complex SQL queries and discover new insights. They can also investigate more on the multi-step instruction prompting~(Section~\ref{app:multi_step}) and chain-of-thought prompting~(Section~\ref{app:chain_of_thought}) to better understand LLMs.

\subsection{Input Format Analysis}
\label{app:input}

In this section, we focus on comparing two formats of inputting tables, namely \textit{markdown} and \textit{flatten}, to explore their impact on LLMs performance. Figure~\ref{fig:format} clearly demonstrates a significant improvement in the model's performance when the \textit{flatten} format is used instead of the \textit{markdown} input format at any experiments settings.

The reason behind this improvement lies in the structure of the SQL template, specifically ``select \textless{}col1\textgreater{} where \textless{}col2\textgreater{} \textless{}op\textgreater{} \textless{}int2\textgreater{}''. In order to execute this template, the model needs to locate the column corresponding to col2 and then identify the row where ``int2'' is found. This process involves 2-hop reasoning. In \textit{markdown} mode, the challenge lies not only in the LLM's understanding of the table structure but also in how to navigate to another column in the same row. However, in \textit{flatten} mode, redundant columns are added to each row as ``Column is value.'' This additional information simplifies the LLM's understanding of the table structure and facilitates reasoning. As a result, the flatten method proves to be more beneficial for LLM performance due to its enhanced structure comprehension and reasoning capabilities.

\subsection{SQL Keywords Analysis}
\label{app:sql_keywords}

\begin{figure*}[t]
  \centering
    \begin{minipage}[b]{0.45\textwidth}
        \centering
    \includegraphics[width=\textwidth]{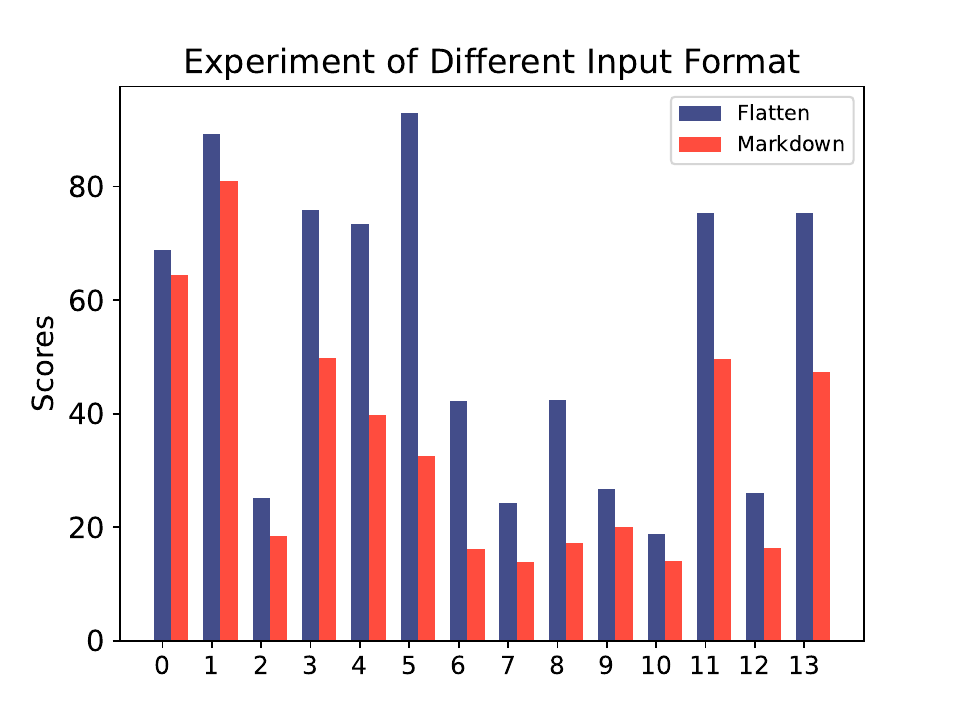}
    \caption{Different input format.}
    \label{fig:format}
  \end{minipage}
  \hfill
  \begin{minipage}[b]{0.45\textwidth}
    \centering
    \includegraphics[width=\textwidth]{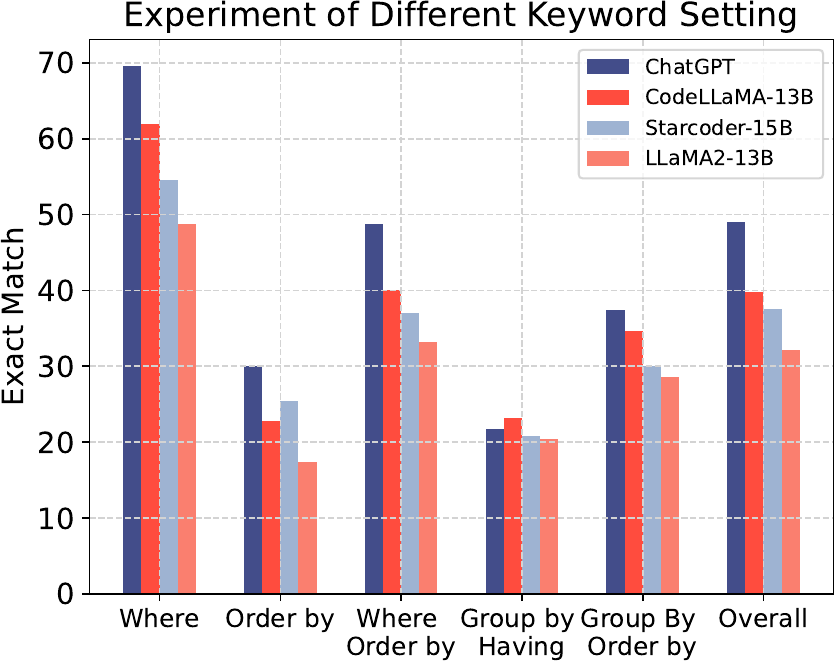}
    \caption{Different keywords setting.}
    \label{fig:keyword}
  \end{minipage}
\end{figure*}

SQL statements follow a specific syntax and are a well-established language in the database domain. We first control SQL statements to contain only specific types of keywords from the perspective of SQL keywords and test the performance of different models on \eval. The experimental results are shown in Figure~\ref{fig:keyword}. The change in the performance of LLMs on SQL statements reflects the trend in the difficulty of reasoning.

\subsection{SQL Attribute Analysis}
\label{app:sql_attribute}

\begin{figure*}[tb]
\centering
\begin{subfigure}[b]{0.32\textwidth} 
  \centering
  {\includegraphics[width=\textwidth]{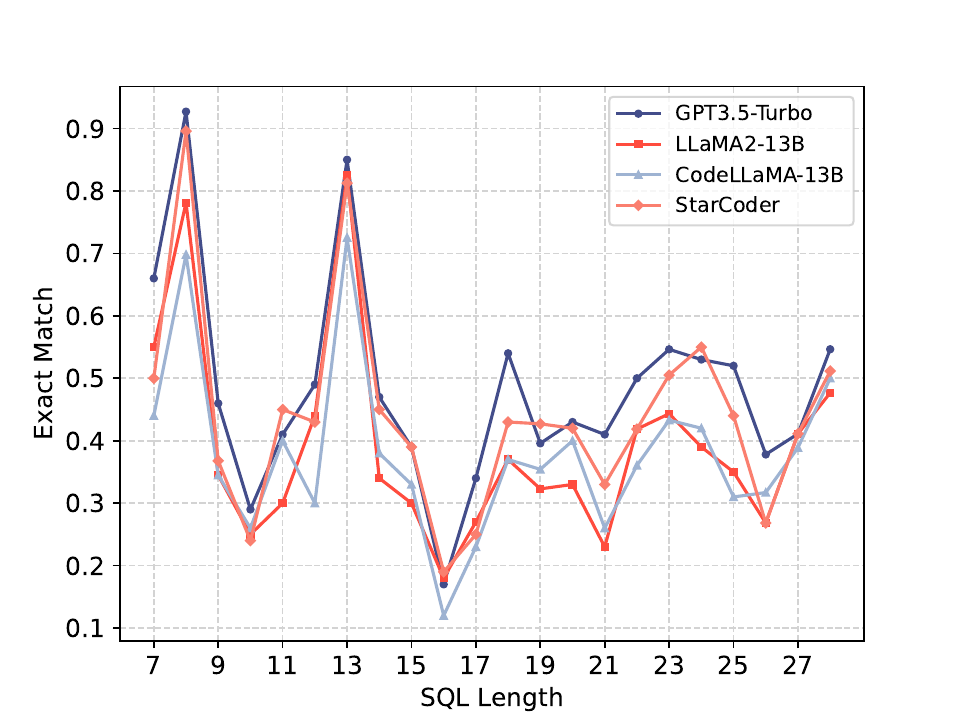}}
  \caption{SQL length}\label{fig:length}
\end{subfigure}
\hfill
\begin{subfigure}[b]{0.32\textwidth} 
  \centering
  {\includegraphics[width=\textwidth]{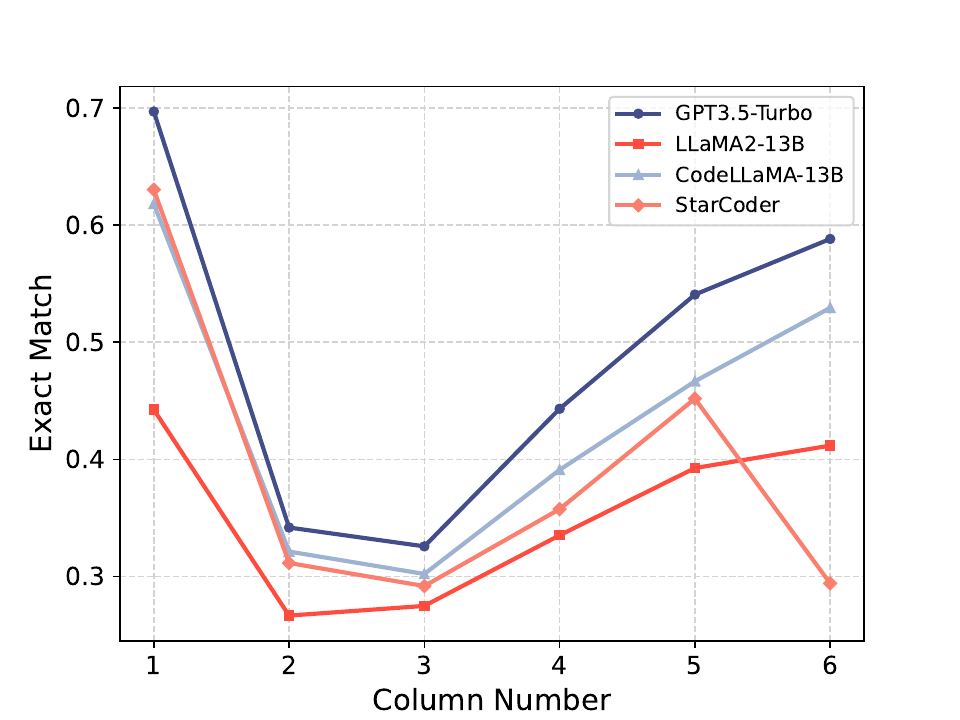}}
  \caption{Column number.}\label{fig:column}
\end{subfigure}
\hfill
\begin{subfigure}[b]{0.32\textwidth} 
  \centering
  {\includegraphics[width=\textwidth]{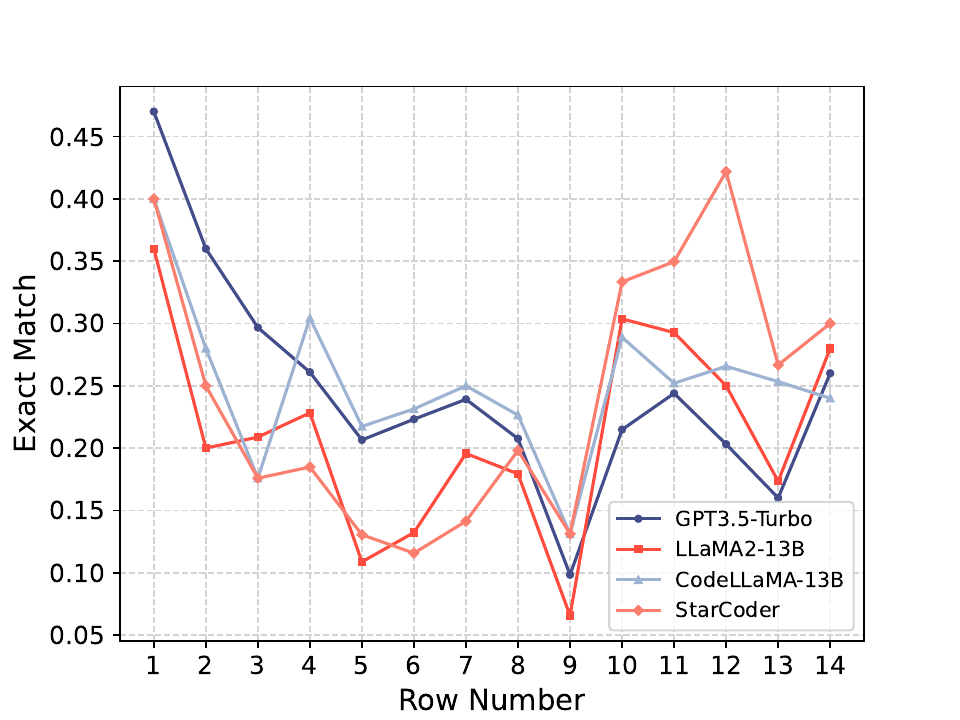}}
  \caption{Selected row number}\label{fig:row}
\end{subfigure}

\begin{subfigure}[b]{0.32\textwidth} 
  \centering
  {\includegraphics[width=\textwidth]{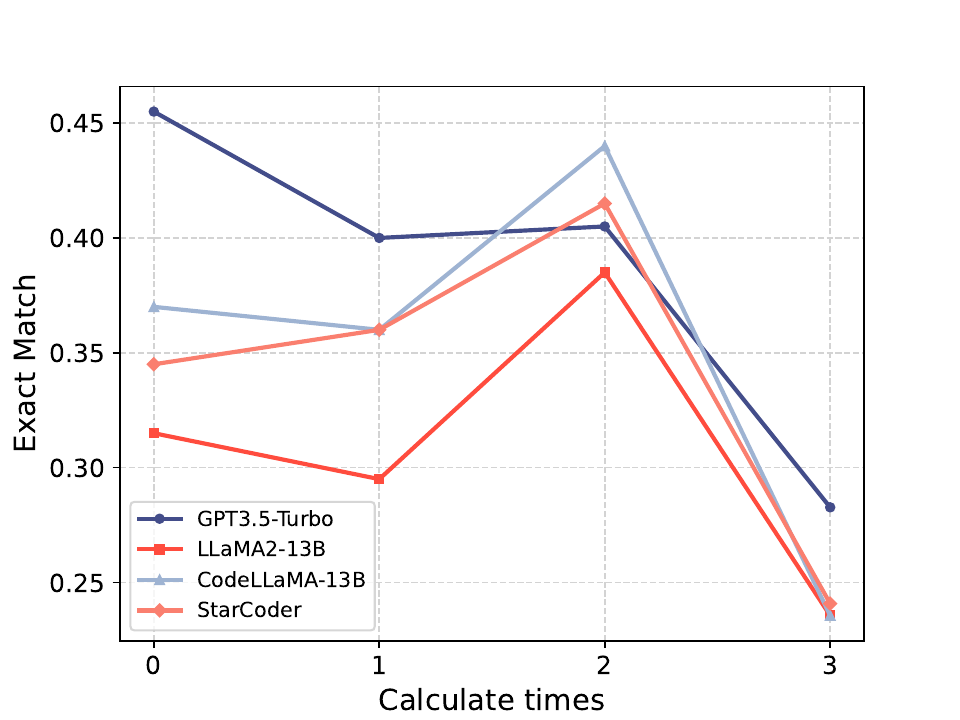}}
  \caption{Calculation times}\label{fig:calc}
\end{subfigure}
\hfill
\begin{subfigure}[b]{0.32\textwidth} 
  \centering
  {\includegraphics[width=\textwidth]{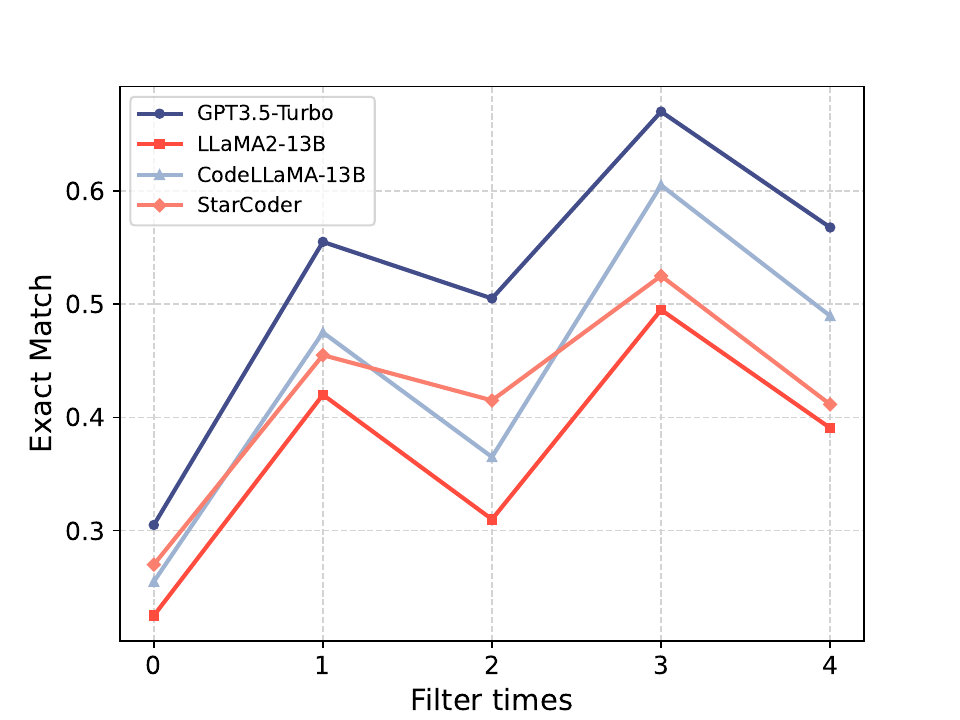}}
  \caption{Filter times.}\label{fig:filter}
\end{subfigure}
\hfill
\begin{subfigure}[b]{0.32\textwidth} 
  \centering
  {\includegraphics[width=\textwidth]{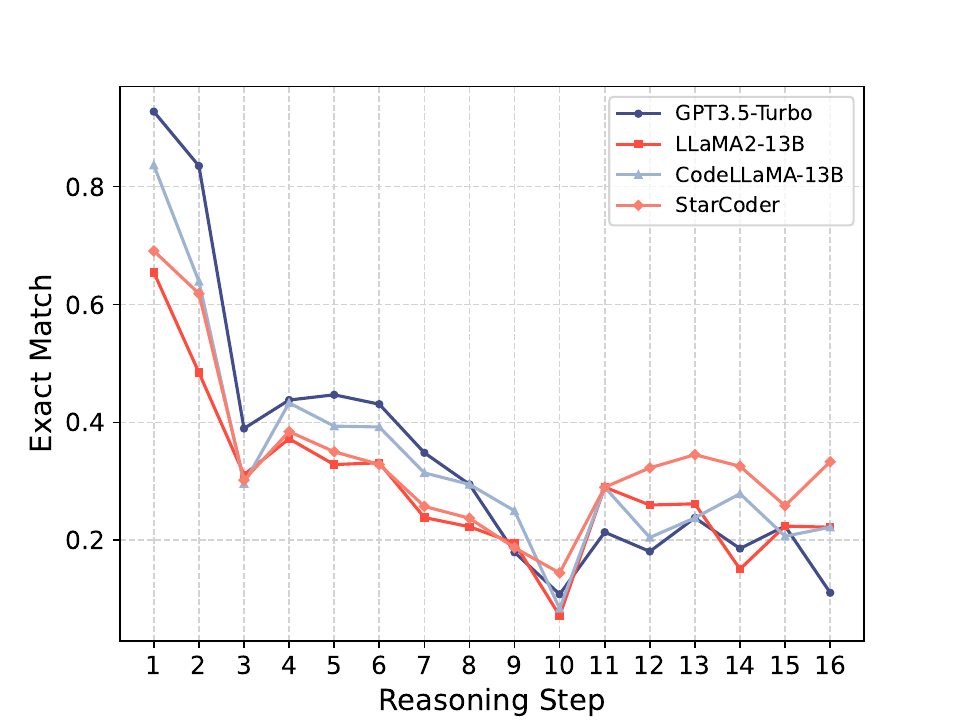}}
  \caption{Total analysis.}\label{fig:total}
\end{subfigure}

\caption{Effect of SQL Attribute Settings on model performance.}
\label{fig:length_column}
\end{figure*}

\eval has the ability to flexibly modify the properties of generated SQL statements, including the length of the statement, the number of computations, and the quantity of filtering numbers. These features can intuitively impact the complexity of SQL. In our experiments, we set the table size to $15\times10$ and adjusted the SQL settings for examining the effect of different SQL attributes on model performance. For example, in the analysis of "Calculation Times," we employed 500 samples with 0, 1, 2, and 3 calculation times respectively. The experimental outcomes of all SQL attributes are illustrated in Figure~\ref{fig:length_column}. While it might be expected that model performance would decline as these factor values increase, the performance actually fluctuates. Upon combining Column number, Row number, Calculation times, and Filter times in the statistical analysis, we identified a significant downward trend in the model, as demonstrated in Figure~\ref{fig:total}.

\label{app:answer_position}

\onecolumn

\section{Data Demostration}

\subsection{Dense and Sparse Examples}
\label{app:neighbor_disperse}

SQL: select boarfish from w where sixties = 'jcrbb'

Answer: ['qxgd', 'lorfaljob', 'qytocp', 'vkfzhqwj', 'xwijyubr']

We can find that \textit{Dense} Setting is better than \textit{Sparse} Setting in all cases.

\paragraph{Sparse Example:}
{\small
\begin{Verbatim}[commandchars=\\\{\}]
|    | boarfish   | tool     | sixties   | phoxinus   | angling   |                  
|---:|:-----------|:---------|:----------|:-----------|:----------|
|  0 | mjdsv      | cwzqkdte | tbwqa     | yuogpbo    | mkxqnrhq  |
|  1 | nrbmyc     | eqciiims | wvfesrtzt | yvvgzj     | mkxqnrhq  |
|  2 | iqdr       | ezhuj    | bndktpe   | yuogpbo    | yjblg     |
|  3 | \textcolor{red}{qxgd}       | dtfjqfc  | \textcolor{purple}{jcrbb}     | haxyaz     | yjblg     |
|  4 | xzrrs      | ezhuj    | bndktpe   | dpimlb     | skbpzyhak |
|  5 | \textcolor{red}{lorfaljob}  | eqciiims | \textcolor{purple}{jcrbb}     | jsvbugac   | bwxihx    |
|  6 | pvugxgdju  | dtfjqfc  | bndktpe   | jsvbugac   | mkxqnrhq  |
|  7 | xpkuautv   | ezhuj    | vyoo      | yvvgzj     | bwxihx    |
|  8 | afzrom     | jzdra    | bndktpe   | jsvbugac   | mkxqnrhq  |
|  9 | ivxpmv     | eqciiims | bndktpe   | jsvbugac   | bwxihx    |
| 10 | ehfvur     | ezhuj    | tbwqa     | yuogpbo    | bwxihx    |
| 11 | bdzsy      | ezhuj    | bndktpe   | yvvgzj     | yjblg     |
| 12 | qruh       | ezhuj    | bndktpe   | dpimlb     | skbpzyhak |
| 13 | \textcolor{red}{qytocp}     | jzdra    | \textcolor{purple}{jcrbb}     | dpimlb     | bwxihx    |
| 14 | eqaja      | ezhuj    | bndktpe   | haxyaz     | yjblg     |
| 15 | kwvzixe    | jzdra    | vyoo      | jsvbugac   | skbpzyhak |
| 16 | edmkxm     | eqciiims | vyoo      | haxyaz     | mkxqnrhq  |
| 17 | fdsdlcpxj  | eqciiims | vyoo      | dpimlb     | blqoislm  |
| 18 | ipprxzzlv  | cwzqkdte | bndktpe   | yuogpbo    | yjblg     |
| 19 | gqyxjtbz   | eqciiims | tbwqa     | dpimlb     | yjblg     |
| 20 | noqfw      | ezhuj    | vyoo      | haxyaz     | blqoislm  |
| 21 | \textcolor{red}{vkfzhqwj}   | dtfjqfc  | \textcolor{purple}{jcrbb}     | yuogpbo    | mkxqnrhq  |
| 22 | konftq     | eqciiims | vyoo      | dpimlb     | bwxihx    |
| 23 | ymcwhu     | jzdra    | wvfesrtzt | dpimlb     | blqoislm  |
| 24 | kpygsu     | eqciiims | wvfesrtzt | yuogpbo    | yjblg     |
| 25 | tiwfvqgmt  | ezhuj    | bndktpe   | dpimlb     | mkxqnrhq  |
| 26 | ovomhf     | dtfjqfc  | bndktpe   | yuogpbo    | blqoislm  |
| 27 | lokwxn     | cwzqkdte | tbwqa     | yuogpbo    | mkxqnrhq  |
| 28 | \textcolor{red}{xwijyubr}   | jzdra    | \textcolor{purple}{jcrbb}     | yuogpbo    | mkxqnrhq  |
| 29 | ttonww     | dtfjqfc  | wvfesrtzt | haxyaz     | blqoislm  |
\end{Verbatim}
}

\paragraph{Dense Example:}
{\small
\begin{Verbatim}[commandchars=\\\{\}]
|    | boarfish   | tool     | sixties   | phoxinus   | angling   |
|---:|:-----------|:---------|:----------|:-----------|:----------|
|  0 | mjdsv      | cwzqkdte | tbwqa     | yuogpbo    | mkxqnrhq  |
|  1 | nrbmyc     | eqciiims | wvfesrtzt | yvvgzj     | mkxqnrhq  |
|  2 | iqdr       | ezhuj    | bndktpe   | yuogpbo    | yjblg     |
|  3 | xzrrs      | ezhuj    | bndktpe   | dpimlb     | skbpzyhak |
|  4 | pvugxgdju  | dtfjqfc  | bndktpe   | jsvbugac   | mkxqnrhq  |
|  5 | xpkuautv   | ezhuj    | vyoo      | yvvgzj     | bwxihx    |
|  6 | afzrom     | jzdra    | bndktpe   | jsvbugac   | mkxqnrhq  |
|  7 | ivxpmv     | eqciiims | bndktpe   | jsvbugac   | bwxihx    |
|  8 | ehfvur     | ezhuj    | tbwqa     | yuogpbo    | bwxihx    |
|  9 | bdzsy      | ezhuj    | bndktpe   | yvvgzj     | yjblg     |
| 10 | qruh       | ezhuj    | bndktpe   | dpimlb     | skbpzyhak |
| 11 | eqaja      | ezhuj    | bndktpe   | haxyaz     | yjblg     |
| 12 | kwvzixe    | jzdra    | vyoo      | jsvbugac   | skbpzyhak |
| 13 | \textcolor{red}{qxgd}       | dtfjqfc  | \textcolor{purple}{jcrbb}     | haxyaz     | yjblg     |
| 14 | \textcolor{red}{lorfaljob}  | eqciiims | \textcolor{purple}{jcrbb}     | jsvbugac   | bwxihx    |
| 15 | \textcolor{red}{qytocp}     | jzdra    | \textcolor{purple}{jcrbb}     | dpimlb     | bwxihx    |
| 16 | \textcolor{red}{vkfzhqwj}   | dtfjqfc  | \textcolor{purple}{jcrbb}     | yuogpbo    | mkxqnrhq  |
| 17 | \textcolor{red}{xwijyubr}   | jzdra    | \textcolor{purple}{jcrbb}     | yuogpbo    | mkxqnrhq  |
| 18 | edmkxm     | eqciiims | vyoo      | haxyaz     | mkxqnrhq  |
| 19 | fdsdlcpxj  | eqciiims | vyoo      | dpimlb     | blqoislm  |
| 20 | ipprxzzlv  | cwzqkdte | bndktpe   | yuogpbo    | yjblg     |
| 21 | gqyxjtbz   | eqciiims | tbwqa     | dpimlb     | yjblg     |
| 22 | noqfw      | ezhuj    | vyoo      | haxyaz     | blqoislm  |
| 23 | konftq     | eqciiims | vyoo      | dpimlb     | bwxihx    |
| 24 | ymcwhu     | jzdra    | wvfesrtzt | dpimlb     | blqoislm  |
| 25 | kpygsu     | eqciiims | wvfesrtzt | yuogpbo    | yjblg     |
| 26 | tiwfvqgmt  | ezhuj    | bndktpe   | dpimlb     | mkxqnrhq  |
| 27 | ovomhf     | dtfjqfc  | bndktpe   | yuogpbo    | blqoislm  |
| 28 | lokwxn     | cwzqkdte | tbwqa     | yuogpbo    | mkxqnrhq  |
| 29 | ttonww     | dtfjqfc  | wvfesrtzt | haxyaz     | blqoislm  |
\end{Verbatim}
}

\subsection{SQL Template}
\label{app:template}

\paragraph{General:}
{\small
\begin{verbatim}
select <select_condition> from my_table

select <select_condition> from my_table <where_condition>

select <select_condition> from my_table <order_condition>,

select <select_condition> from my_table <where_condition> <order_condition>,

select <select_condition> from my_table <group_condition> <having_condition>,

select <select_condition> from my_table <where_condition> <group_condition> <having_condition>,

select <select_condition> from my_table <where_condition> 
<group_condition> <having_condition> <order_condition>,

select <select_condition> from my_table <group_condition> <having_condition> <order_condition>
\end{verbatim}
}

\paragraph{Where Condition:}
{\small
\begin{verbatim}
select <text_col1> from my_table where <text_col2> = <text_2>
\end{verbatim}
}

\paragraph{Count:}
{\small
\begin{verbatim}
Select Count(<text_col1>) from table where <text_col1> = <text_1>
\end{verbatim}
}

\paragraph{Easy:}
{\small
\begin{verbatim}
select <text_col1> from my_table where <int_col1> = <int_1>
select <int_col1> from my_table where <text_col1> = <text_1>
select <int_col1> from my_table where <int_col2> = <int_2>
select <text_col1> from my_table where <text_col2> = <text_2>
\end{verbatim}
}

\paragraph{Filter:}
{\small
\begin{verbatim}
select <text_col1> from my_table where <text_col2> = <text_2>
select <text_col1> from my_table where <int_col2> <op2> <int_2>
select <text_col1> from my_table where <text_col2> = <text_2> and <int_col1> <op1> <int_1>
select <text_col1> from my_table where <text_col2> = <text_2> and <text_col3> = <text_3>
select <text_col1> from my_table where <int_col1> <op1> <int_1> and <int_col2> <op2> <int_2>
select <int_col1> from my_table where <text_col1> = <text_1>
select <int_col1> from my_table where <int_col2> <op2> <int_2>
select <int_col1> from my_table where <text_col2> = <text_2> and <int_col2> <op2> <int_2>
select <int_col1> from my_table where <text_col2> = <text_2> and <text_col3> = <text_3>
select <int_col1> from my_table where <int_col2> <op2> <int_2> and <int_col3> <op3> <int_3>

\end{verbatim}
}

\paragraph{Aggregate:}
{\small
\begin{verbatim}
select count ( <text_col1> ) from my_table where <text_col2> = <text_2>
select count ( <text_col1> ) from my_table where <int_col2> <op2> <int_2>
select sum ( <int_col1> ) from my_table
select sum ( <int_col1> ) from my_table where <text_col2> = <text_2>
select max ( <int_col1> ) from my_table
select max ( <int_col1> ) from my_table where <text_col2> = <text_2>
select min ( <int_col1> ) from my_table
select min ( <int_col1> ) from my_table where <text_col2> = <text_2>
\end{verbatim}
}

\paragraph{Arithmetic:}
{\small
\begin{verbatim}
select <int_col1> + <int_col2> from my_table where <text_col1> = <text_1>
select <int_col1> + <int_col2> from my_table where <text_col1> = <text_1> and <text_col2> = <text_2>
select <int_col1> - <int_col2> from my_table where <text_col1> = <text_1>
select <int_col1> - <int_col2> from my_table where <text_col1> = <text_1> and <text_col2> = <text_2>
\end{verbatim}
}

\paragraph{Superlative:}
{\small
\begin{verbatim}
select <int_col1> from my_table order by <int_col1> asc limit 1
select <int_col1> from my_table order by <int_col1> desc limit 1
select <text_col1> from my_table order by <int_col1> asc limit 1
select <text_col1> from my_table order by <int_col1> desc limit 1
select <int_col1> from my_table order by <int_col2> asc limit 1
select <int_col1> from my_table order by <int_col2> desc limit 1
\end{verbatim}
}

\paragraph{Comparative:}
{\small
\begin{verbatim}
select ( select <int_col1> from my_table where <text_col1> = <text_1> ) 
> ( select <int_col1> from my_table where <text_col2> = <text_2> )
select ( select <int_col1> from my_table where <int_col2> <op2> <int_2> ) 
> ( select <int_col1> from my_table where <int_col3> <op3> <int_3> )
select ( select <int_col1> from my_table where <text_col1> = <text_1> ) 
< ( select <int_col1> from my_table where <text_col2> = <text_2> )
select ( select <int_col1> from my_table where <int_col2> <op2> <int_2> ) 
< ( select <int_col1> from my_table where <int_col3> <op3> <int_3> )
select <int_col1> > <int_col2> from my_table where <text_col1> = <text_1>
select <int_col1> < <int_col2> from my_table where <text_col1> = <text_1>
select <int_col1> > <int_col2> from my_table where <int_col3> <op3> <int_3>
select <int_col1> < <int_col2> from my_table where <int_col3> <op3> <int_3>
\end{verbatim}
}

\subsection{Table Input Format}
\label{app:table_input_format}

\paragraph{Markdown Table:}
{\small
\begin{verbatim}
|    |   ercilla |   shucks | liter    | taenia    | dorado    |
|---:|----------:|---------:|:---------|:----------|:----------|
|  0 |        68 |       12 | gcrdvo   | qoath     | katfuw    |
|  1 |       129 |      151 | zmvltkk  | jpcglcjzk | vwqqey    |
|  2 |       248 |      188 | zmdlfbhb | cvhqotsys | wzunmaa   |
|  3 |       267 |      104 | gcrdvo   | ytywunvf  | pjlbo     |
|  4 |       135 |      262 | gcrdvo   | dtnvfp    | ajzpsaoy  |
|  5 |       309 |      119 | zmdlfbhb | klcenmugk | hriunhf   |
|  6 |        25 |      152 | zmvltkk  | cjgcergv  | shrbvrd   |
|  7 |       298 |       18 | zmvltkk  | scvuuc    | ahunvcx   |
|  8 |       321 |      217 | gcrdvo   | ezlp      | hasjaznm  |
|  9 |       139 |      310 | gcrdvo   | ghhjea    | atqvtgoa  |
| 10 |        99 |       34 | zmvltkk  | ecdmpruq  | cfitvz    |
| 11 |       142 |      167 | gcrdvo   | acii      | oenmuezip |
| 12 |       273 |      156 | gcrdvo   | nnvnteh   | tulh      |
| 13 |       197 |       44 | gcrdvo   | pqdbhevkh | dfxuwxz   |
| 14 |       144 |      123 | gcrdvo   | bxrgo     | ccbj      |
\end{verbatim}
}

\paragraph{Flatten Table:}
{\small
\begin{verbatim}
Flatten Table Examples:
The table have 5 columns: ercilla | shucks | liter | taenia | dorado
row 1 : ercilla is 68. shucks is 12. liter is gcrdvo. taenia is qoath. dorado is katfuw. 
row 2 : ercilla is 129. shucks is 151. liter is zmvltkk. taenia is jpcglcjzk. dorado is vwqqey. 
row 3 : ercilla is 248. shucks is 188. liter is zmdlfbhb. taenia is cvhqotsys. dorado is wzunmaa. 
row 4 : ercilla is 267. shucks is 104. liter is gcrdvo. taenia is ytywunvf. dorado is pjlbo. 
row 5 : ercilla is 135. shucks is 262. liter is gcrdvo. taenia is dtnvfp. dorado is ajzpsaoy. 
row 6 : ercilla is 309. shucks is 119. liter is zmdlfbhb. taenia is klcenmugk. dorado is hriunhf. 
row 7 : ercilla is 25. shucks is 152. liter is zmvltkk. taenia is cjgcergv. dorado is shrbvrd. 
row 8 : ercilla is 298. shucks is 18. liter is zmvltkk. taenia is scvuuc. dorado is ahunvcx. 
row 9 : ercilla is 321. shucks is 217. liter is gcrdvo. taenia is ezlp. dorado is hasjaznm. 
row 10 : ercilla is 139. shucks is 310. liter is gcrdvo. taenia is ghhjea. dorado is atqvtgoa. 
row 11 : ercilla is 99. shucks is 34. liter is zmvltkk. taenia is ecdmpruq. dorado is cfitvz. 
row 12 : ercilla is 142. shucks is 167. liter is gcrdvo. taenia is acii. dorado is oenmuezip. 
row 13 : ercilla is 273. shucks is 156. liter is gcrdvo. taenia is nnvnteh. dorado is tulh. 
row 14 : ercilla is 197. shucks is 44. liter is gcrdvo. taenia is pqdbhevkh. dorado is dfxuwxz. 
row 15 : ercilla is 144. shucks is 123. liter is gcrdvo. taenia is bxrgo. dorado is ccbj. 
\end{verbatim}
}

\subsection{SQL Execution Examples~(Few-shot)}
\label{app:multi_step}
\paragraph{}
{\small
\begin{verbatim}
You are an SQL executor, you need to execute SQL based on the give table and SQL statement 
to obtain the execution results. 
Only give me the execution results and do not output any other words. 
Table:
|    |   puccoon |   tiepolo |   scope | mutinus    |   intrados | huggins   |   barye |   wear |
|---:|----------:|----------:|--------:|:-----------|-----------:|:----------|--------:|-------:|
|  0 |       171 |       225 |     145 | 2007-04-27 |        322 | yefihroyn |      79 |    207 |
|  1 |       213 |       116 |     319 | 2016-01-15 |        288 | ytyayrvj  |     246 |    272 |
|  2 |       191 |       229 |      95 | 2022-11-08 |        218 | gpmvax    |     167 |     73 |
|  3 |        97 |       155 |     189 | 2013-10-30 |         79 | gpmvax    |      24 |    233 |
|  4 |        56 |        11 |     295 | 2018-12-10 |         81 | yefihroyn |     187 |    198 |
|  5 |       285 |       304 |     168 | 2017-03-24 |         75 | gpmvax    |     111 |     77 |
|  6 |       233 |       325 |      31 | 2014-01-22 |        114 | ytyayrvj  |      20 |    219 |
|  7 |        19 |       146 |     164 | 2021-12-07 |        311 | ytyayrvj  |     188 |      3 |
|  8 |       112 |       255 |      30 | 2015-12-07 |        214 | gpmvax    |      16 |    271 |
|  9 |       175 |        62 |     181 | 2012-04-21 |        182 | gpmvax    |     105 |     76 |
| 10 |       200 |        90 |     101 | 2008-04-28 |        168 | gpmvax    |      70 |    119 |
| 11 |        31 |       180 |      95 | 2004-06-23 |         62 | yefihroyn |     314 |     97 |
| 12 |       297 |       251 |     249 | 2022-02-02 |        185 | yefihroyn |     278 |    313 |
| 13 |        36 |        17 |      67 | 2016-04-14 |        243 | ytyayrvj  |     213 |      4 |
| 14 |        45 |       215 |     182 | 2012-06-15 |        251 | yefihroyn |     221 |     83 |
Now you need to execute SQL based on the given table and SQL statement to obtain the execution result. 
Only give me the result and do not output any other words or SQL statement.
The following are some examples.

SQL:select avg ( intrados ) from my_table where tiepolo > 146 group by huggins 
having count ( huggins ) > 1 order by count ( tiepolo ) asc limit 1
Answer:146.5
SQL:select wear from my_table where huggins = 'gpmvax' group by huggins 
having wear < 83 order by count ( distinct barye ) asc limit 1
Answer:73
SQL:select mutinus from my_table where tiepolo > 116 group by huggins 
having max ( wear ) > 119 order by count ( huggins ) asc limit 1
Answer:2014-01-22
SQL:select tiepolo from my_table where puccoon < 191 and intrados < 79 group by huggins 
having intrados < 81 and tiepolo < 255 order by count ( barye ) asc limit 1
Answer:180
SQL:select tiepolo from my_table where scope > 31 group by huggins 
having min ( tiepolo ) = 62 order by count ( distinct mutinus ) asc limit 1
Answer:62
SQL:select wear from my_table where huggins = 'ytyayrvj' group by huggins 
having count ( huggins ) < 5 order by count ( distinct mutinus ) desc limit 1
Answer:
\end{verbatim}
}

\subsection{SQL Execution Examples (Multi-Answer)}
\label{app:multi_answer}
\paragraph{}
{\small
\begin{verbatim}

You are an SQL executor, you need to execute SQL based on the give table 
and SQL statement to obtain the execution results.
| suiting   | chisel    |   highboy |   broccoli | newburgh   | acetum    |   brewpub |
|:----------|:----------|----------:|-----------:|:-----------|:----------|----------:|
| zbwamhiui | nnkfvevxw |        50 |         88 | zhwohj     | opufj     |       214 |
| zroosgm   | yvftt     |       309 |        168 | zhwohj     | xqsu      |       136 |
| zroosgm   | lnri      |       152 |         78 | zhwohj     | ikvsd     |       219 |
| kjsdl     | trei      |       234 |        287 | egkgkvbec  | mhxcxyg   |        23 |
| zroosgm   | mctnpwbd  |        71 |        242 | egkgkvbec  | yszfokeom |       180 |
| zbwamhiui | ptqtj     |        19 |         81 | egkgkvbec  | hyfmk     |       116 |
| zroosgm   | lpjvwn    |       258 |        313 | uftnwbd    | oevmj     |        65 |
| kjsdl     | ididumrhw |        64 |        101 | uftnwbd    | xjakwpayx |       327 |
| zbwamhiui | wdtncbyn  |       165 |        209 | uftnwbd    | xrbqvxb   |       192 |
| zbwamhiui | wyjjc     |       219 |          6 | uftnwbd    | pzqr      |       188 |
| zroosgm   | qumxgwvls |       314 |        246 | uftnwbd    | ehevtf    |        60 |
| zbwamhiui | adiyf     |       207 |        298 | egkgkvbec  | wbrgejgf  |        80 |
| zbwamhiui | qpgpbj    |       307 |        306 | egkgkvbec  | mcjuonhc  |       192 |
| zbwamhiui | ehsk      |        47 |        244 | zhwohj     | tcdlnc    |       280 |
| kjsdl     | orlosbok  |        21 |         93 | egkgkvbec  | dzvwohjo  |       103 |
| zbwamhiui | webyyylw  |        84 |        195 | egkgkvbec  | xbmv      |       289 |
| kjsdl     | mrcecp    |        48 |        264 | egkgkvbec  | xhprcocik |       265 |
| kjsdl     | ngajupd   |       247 |         52 | zhwohj     | pcokyw    |       247 |
| zroosgm   | xeeuixkze |       120 |        288 | zhwohj     | yishnriw  |       138 |
| kjsdl     | kbczy     |       119 |         13 | egkgkvbec  | ltpmyfdt  |        73 |
| zbwamhiui | uvvdzo    |       150 |         57 | uftnwbd    | tajlsm    |       295 |
| zbwamhiui | enbffevhp |       290 |         92 | zhwohj     | gjjznp    |        18 |
| zroosgm   | imubtcc   |        79 |         19 | uftnwbd    | eqymwj    |       112 |

SQL:select suiting from my_table group by suiting having count ( newburgh ) > 6
Answer:
| suiting   |
|:----------|
| zbwamhiui |
| zroosgm   |

SQL:select acetum,newburgh,suiting from my_table where highboy > 234
Answer:
| acetum   | newburgh   | suiting   |
|:---------|:-----------|:----------|
| xqsu     | zhwohj     | zroosgm   |
| oevmj    | uftnwbd    | zroosgm   |
| ehevtf   | uftnwbd    | zroosgm   |
| mcjuonhc | egkgkvbec  | zbwamhiui |
| pcokyw   | zhwohj     | kjsdl     |
| gjjznp   | zhwohj     | zbwamhiui |

SQL:select count ( chisel ) from my_table where highboy < brewpub 
group by newburgh having min ( highboy ) < 47 
Answer:
|   count ( chisel ) |
|-------------------:|
|                  5 |

SQL:select newburgh from my_table where brewpub > 138 order by broccoli desc limit 1
Answer:
| newburgh   |
|:-----------|
| egkgkvbec  |


SQL:select suiting from my_table where highboy > broccoli 
group by suiting having min ( highboy ) < 314

Answer:

\end{verbatim}
}

\subsection{Multi-step Instruction~(Few-shot)}
\label{app:multi_step}
\paragraph{}
{\small
\begin{verbatim}
You need to obtain the final answer based on the table and instructions. 
Only give me the result and do not output any other words. 
Table:
|    |   puccoon |   tiepolo |   scope | mutinus    |   intrados | huggins   |   barye |   wear |
|---:|----------:|----------:|--------:|:-----------|-----------:|:----------|--------:|-------:|
|  0 |       171 |       225 |     145 | 2007-04-27 |        322 | yefihroyn |      79 |    207 |
|  1 |       213 |       116 |     319 | 2016-01-15 |        288 | ytyayrvj  |     246 |    272 |
|  2 |       191 |       229 |      95 | 2022-11-08 |        218 | gpmvax    |     167 |     73 |
|  3 |        97 |       155 |     189 | 2013-10-30 |         79 | gpmvax    |      24 |    233 |
|  4 |        56 |        11 |     295 | 2018-12-10 |         81 | yefihroyn |     187 |    198 |
|  5 |       285 |       304 |     168 | 2017-03-24 |         75 | gpmvax    |     111 |     77 |
|  6 |       233 |       325 |      31 | 2014-01-22 |        114 | ytyayrvj  |      20 |    219 |
|  7 |        19 |       146 |     164 | 2021-12-07 |        311 | ytyayrvj  |     188 |      3 |
|  8 |       112 |       255 |      30 | 2015-12-07 |        214 | gpmvax    |      16 |    271 |
|  9 |       175 |        62 |     181 | 2012-04-21 |        182 | gpmvax    |     105 |     76 |
| 10 |       200 |        90 |     101 | 2008-04-28 |        168 | gpmvax    |      70 |    119 |
| 11 |        31 |       180 |      95 | 2004-06-23 |         62 | yefihroyn |     314 |     97 |
| 12 |       297 |       251 |     249 | 2022-02-02 |        185 | yefihroyn |     278 |    313 |
| 13 |        36 |        17 |      67 | 2016-04-14 |        243 | ytyayrvj  |     213 |      4 |
| 14 |        45 |       215 |     182 | 2012-06-15 |        251 | yefihroyn |     221 |     83 |
Now you need to get the answer based on the instruction, 
only give me the result and do not output any other words.
The following are some examples.

Instruction:Please filter the rows by the column conditions, which need to be met: 
The value of column tiepolo needs to be greater than 146.
The rows are then grouped according to the value of the huggins in the remaining rows.
Then filter some groups by the following condition:the number of column huggins is greater than 1.
Select the average of values of intrados column in filtered rows.
Sort the obtained values in ascending order of the number of tiepolo 
and select the smallest value to get the answer.
Answer:146.5

Instruction:Please filter the rows by the column conditions, which need to be met: 
The value of column huggins is 'gpmvax'.
The rows are then grouped according to the value of the huggins in the remaining rows.
Then filter some groups by the following condition:the column wear is less than 83.
Select values of wear column in filtered rows.
Sort the obtained values in ascending order of the number of non-repeating barye 
and select the smallest value to get the answer.
Answer:73

Instruction:Please filter the rows by the column conditions, which need to be met: 
The value of column huggins is 'ytyayrvj'.
The rows are then grouped according to the value of the huggins in the remaining rows.
Then filter some groups by the following condition:the number of column huggins is less than 5.
Select values of wear column in filtered rows.
Sort the obtained values in descending order of the number of non-repeating mutinus 
and select the largest value to get the answer.
Answer:
\end{verbatim}
}

\subsection{Chain-of-Thought SQL Execution Prompting Examples}
\label{app:chain_of_thought}

{\small
\begin{verbatim}
You are an SQL executor, you need to output the execution process and final answer based on table and SQL.
Table:
|    | masthead   | laertes   |   boo | bothrops   | height   | scraper   | trouser   |   lozenge |
|---:|:-----------|:----------|------:|:-----------|:---------|:----------|:----------|----------:|
|  0 | case       | araeswrid |    41 | lyucg      | urbsmxiv | vgxrh     | esauw     |       281 |
|  1 | case       | araeswrid |   138 | lyucg      | tbvg     | oerigocb  | stevw     |       177 |
|  2 | case       | zncmrrvg  |   303 | loclzoglg  | tbvg     | vgxrh     | stevw     |       234 |
|  3 | thyngfwts  | araeswrid |   288 | loclzoglg  | tbvg     | vgxrh     | esauw     |       224 |
|  4 | thyngfwts  | mrehctv   |   177 | loclzoglg  | urbsmxiv | vgxrh     | esauw     |       228 |
|  5 | case       | araeswrid |   163 | loclzoglg  | urbsmxiv | oerigocb  | stevw     |        60 |
|  6 | thyngfwts  | mrehctv   |    45 | loclzoglg  | cidufm   | oerigocb  | esauw     |       289 |
|  7 | thyngfwts  | zncmrrvg  |    42 | loclzoglg  | tbvg     | ffljyxb   | stevw     |       296 |
|  8 | case       | araeswrid |   275 | lyucg      | cidufm   | vgxrh     | stevw     |       172 |
|  9 | case       | mrehctv   |    20 | loclzoglg  | tbvg     | vgxrh     | esauw     |       147 |
| 10 | thyngfwts  | araeswrid |   302 | lyucg      | urbsmxiv | vgxrh     | stevw     |       297 |
| 11 | thyngfwts  | zncmrrvg  |   137 | loclzoglg  | tbvg     | vgxrh     | esauw     |        63 |
| 12 | case       | araeswrid |   186 | loclzoglg  | cidufm   | ffljyxb   | esauw     |       268 |
| 13 | case       | araeswrid |   194 | loclzoglg  | cidufm   | vgxrh     | esauw     |        98 |
| 14 | case       | araeswrid |   234 | lyucg      | urbsmxiv | vgxrh     | stevw     |       276 |
Now you need to get the answer based on the instruction, 
only give me the intermedium results and the final answer.
SQL:
select masthead from my_table where height = 'tbvg' group by masthead order by count ( laertes ) desc limit 1
Execution process:
You need to execute 3 steps.
Step 0: 
Please filter the rows by the column conditions, which need to be met: The value of column butcher is 'jxys'.
Intermediate results 0:
|    | encyclia   | butcher   | bowdler   | nuthatch   |   cachexia | claret   |   cortina |   strombus |
|---:|:-----------|:----------|:----------|:-----------|-----------:|:---------|----------:|-----------:|
|  0 | adnh       | jxys      | cxjvfz    | clmb       |          2 | oqmdmbfg |       251 |        184 |
|  1 | xvoxfjbm   | jxys      | cxjvfz    | clmb       |        275 | oqmdmbfg |       140 |        303 |
|  2 | adnh       | jxys      | eohdpivo  | clmb       |        298 | oqmdmbfg |       142 |         28 |
|  3 | adnh       | jxys      | eohdpivo  | rcyixdl    |        153 | oqmdmbfg |        50 |        306 |
|  4 | xvoxfjbm   | jxys      | eohdpivo  | rcyixdl    |        315 | rxbttbm  |       201 |         86 |
Step 1: Select values of strombus column in filtered rows.
Intermediate results 1:
184,303,28,306,86
Step 2: Sort the obtained values in ascending order of claret and select the smallest value to get the answer.
Answer: 184
\end{verbatim}
}

\subsection{Real Table SQL Execution (Few-shot)}
{\small
\begin{verbatim}
You are an SQL executor, you need to execute SQL based on the give table 
and SQL statement to obtain the execution results. 
Only give me the execution results and do not output any other words. 
Table:
|    |   id |   agg |   rank | nation             |   gold |   silver |   bronze |   total |
|---:|-----:|------:|-------:|:-------------------|-------:|---------:|---------:|--------:|
|  0 |    1 |     0 |      1 | soviet union       |     50 |       27 |       22 |      99 |
|  1 |    2 |     0 |      2 | united states      |     33 |       31 |       30 |      94 |
|  2 |    3 |     0 |      3 | east germany (gdr) |     20 |       23 |       23 |      66 |
|  3 |    4 |     0 |      4 | west germany (frg) |     13 |       11 |       16 |      40 |
|  4 |    5 |     0 |      5 | japan              |     13 |        8 |        8 |      29 |
|  5 |    6 |     0 |      6 | australia          |      8 |        7 |        2 |      17 |
|  6 |    7 |     0 |      7 | poland             |      7 |        5 |        9 |      21 |
|  7 |    8 |     0 |      8 | hungary            |      6 |       13 |       16 |      35 |
|  8 |    9 |     0 |      9 | bulgaria           |      6 |       10 |        5 |      21 |
|  9 |   10 |     0 |     10 | italy              |      5 |        3 |       10 |      18 |
Now you need to execute SQL based on the given table and SQL statement to obtain the execution result. 
Only give me the result and do not output any other words or SQL statement.
The following are some examples.

SQL:select nation from table where rank = 1
Answer:Soviet Union
SQL:select nation from table where nation != 'bulgaria' 
and total = ( select total from table where nation = 'bulgaria' )
Answer:Poland
SQL:select nation from table order by bronze limit 1
Answer:Australia
SQL:select nation from table order by bronze limit 1
Answer:Australia
SQL:select silver from table order by gold desc limit 1
Answer:
\end{verbatim}
}

\subsection{Real Table Question Answering (Few-shot)}
{\small
\begin{Verbatim}
You need to obtain the final answer based on the table and questions. 
Only give me the answer and do not output any other words. 
Table:
|    |   id |   agg |   rank | nation             |   gold |   silver |   bronze |   total |
|---:|-----:|------:|-------:|:-------------------|-------:|---------:|---------:|--------:|
|  0 |    1 |     0 |      1 | soviet union       |     50 |       27 |       22 |      99 |
|  1 |    2 |     0 |      2 | united states      |     33 |       31 |       30 |      94 |
|  2 |    3 |     0 |      3 | east germany (gdr) |     20 |       23 |       23 |      66 |
|  3 |    4 |     0 |      4 | west germany (frg) |     13 |       11 |       16 |      40 |
|  4 |    5 |     0 |      5 | japan              |     13 |        8 |        8 |      29 |
|  5 |    6 |     0 |      6 | australia          |      8 |        7 |        2 |      17 |
|  6 |    7 |     0 |      7 | poland             |      7 |        5 |        9 |      21 |
|  7 |    8 |     0 |      8 | hungary            |      6 |       13 |       16 |      35 |
|  8 |    9 |     0 |      9 | bulgaria           |      6 |       10 |        5 |      21 |
|  9 |   10 |     0 |     10 | italy              |      5 |        3 |       10 |      18 |
Now you need to get the answer based on the question, 
only give me the answer and do not output any other words.
The following are some examples.

Question:which country was first in rank at the 1972 olympics ?
Answer:Soviet Union
Question:which country won the same amount of medals as bulgaria in these olympics ?
Answer:Poland
Question:which nation won the least number of bronze medals ?
Answer:Australia
Question:which nation received the least bronze medals
Answer:Australia
Question:what number of silver medals was won by the nation with the most gold medals ?
Answer:
\end{Verbatim}
}

\section{Experiments Settings Details}

\subsection{Setting Description}
\label{app:setting_des}

\paragraph{Table Config}
{\small
\begin{verbatim}
"col_min": 5, // the min number of cols
"col_max": 8, // the max number of cols
"row_min": 15,  // the min number of rows
"row_max": 40,  // the max number of rows
"text_int_date": [0.55, 0.35, 0.1], // text,int,date  header ratio
"text_int_date_fix": ["TEXT", "TEXT", "INT", "INT", "INT"], // Specify the type of each header
// Probability of duplicate values in each column
"value_repeat_ratio": [0, 0.2, 0.3, 0, 0, 0, 0, 0, 0.2, 0.5], 
"value_repeat_ratio_fix": ["random", "random"], // Specify the duplicate values of each column
\end{verbatim}
}

\paragraph{SQL Config}
{\small
\begin{verbatim}
"nest": [1],  // Number of SQL nestings. options: [1], [2], [1,2],[1,2, 3]
"keywords_setting": { // if a Keyword is False, then no SQL containing this Keyword is generated.
"select": true,      
"where": true,
"group by": true,
"having": true,
"order by": true
},
"length_setting": {      // control the length of sql
"is_available": false, // To enable this setting, you need to adjust "is_available" to true first.
// 'value' can be set to specific values, such as [13,14,15], 
// if value is null, then the range is used [min, max]
"value": [],
"min": 6,
"max": 16
},
"column_ratio": {        // Controlling the ratio of columns involved in SQL
"is_available": false,  // To enable this setting, you need to adjust "is_available" to true first.
// 'value' can be set to specific values, such as [1,2], Control the number of columns involved in SQL
"value": [],
// if value is null, then the range is used [min, max], it's the used ratio = (used columns) / (all columns)
"min": 0.1,
"max": 0.3
},
"select_row_ratio":{     // Controlling the ratio of rows involved in select keyword
"is_available": false,  // To enable this setting, you need to adjust "is_available" to true first.
// 'value' can be set to specific values, such as [1,2,3,4], Control the number of rows involved in SQL
"value": [],
// if value is null, then the range is used [min, max], it's the used ratio = (select rows) / (all rows)
"min": 0.1,
"max": 0.2
},
// Controlling the calculate times of the sql ['+','-','*','/','sum','count','min','max','avg']
"calculate_times": {
"is_available": false,   // To enable this setting, you need to adjust "is_available" to true first.
"value": [1,2,3,4]      // 'value' can be set to specific values, means the calculate times
},
// Controlling the filter times of the sql ['=','>','<','in','like']
"filter_times": {
"is_available": false,   // To enable this setting, you need to adjust "is_available" to true first.
"value": [1,2,3,4,5]  // 'value' can be set to specific values, means the calculate times
},
// Controlling the location of answer in the table, usually used in long-context understanding
"answer_location": {
"is_available": false,   // To enable this setting, you need to adjust "is_available" to true first.
"value": null,          
"min": 0.1, // if value is null, then the range is used [min, max], 
means that  0.1 <  (Row where answer is located ) / (Row number) < 0.9
"max": 0.9              
},
// usually remains 1 in this repo, we often just test the sql whose answer is from one cell.
"answer_cells_number": 1,
"include": [],
"exclude": [],
"n_shot": 5
\end{verbatim}
}

\subsection{General Setting}
\label{app:general_setting}

\paragraph{Table Config}
{\small
\begin{verbatim}
"col_min": 5,
"col_max": 5,
"row_min": 30,
"row_max": 30,
"text_int_date": [0.5, 0.45, 0.05],
"value_repeat_ratio": [0, 0.2, 0.3, 0, 0, 0, 0, 0, 0, 0.5]
\end{verbatim}
}

\paragraph{SQL Config}
{\small
\begin{verbatim}
  "nest": [1,2,3],
  "select_grammar": [],
  "keywords_setting": { "select": true, 
  "where": true,
    "group by": true,
    "having": true,
    "order by": true
  },
  "length_setting": {
    "is_available": false,
    "value": [],
    "min": 6,
    "max": 16
  },
  "column_ratio": {
    "is_available": false,
    "value": [],
    "min": 0.1,
    "max": 0.3
  },
  "select_row_ratio":{
    "is_available": false,
    "value": [],
    "min": 0,
    "max": 0.2
  },
  "calculate_times": {
    "is_available": false,
    "value": [0]
  },
  "filter_times": {
    "is_available": false,
    "value": [0]
  },
  "answer_location": {
    "is_available": false,
    "row_value": [],
    "column_value":[0],
    "min": 0,
    "max": 1
  },
  "answer_cells_number": 1,
  "multi_test": false,
  "include": [],
  "exclude": [],
  "n_shot": 5
\end{verbatim}
}

\subsection{LLMs Used In This Paper}
\label{app:llm_details}
\textbf{LLMs.} LLaMA2~\citep{touvron2023llama}, Qwen~\citep{bai2023qwen}, InternLM~\citep{2023internlm}, Mistral, XGen~\citep{nijkamp2023xgen}, Falcon~\citep{refinedweb}, phi-1\_5~\citep{textbooks2}, StableLM~\citep{gpt-neox-library}, Pythia~\citep{biderman2023pythia},
CodeLlama~\citep{roziere2023code}, StarCoder~\citep{li2023starcoder}, CodeGen~\citep{Nijkamp2022ACP}.

We all use the official model weight from the Huggingface Models\footnote{\url{https://huggingface.co/models}}. Above we used the model's abbreviation, we list the model's huggingface official label in Table~\ref{tab:appendix_model}.

\begin{table*}[t]
\centering
\begin{tabular}{cc}
\toprule
Model           & Name                             \\ \midrule
Mistral-7B      & mistralai/Mistral-7B-v0.1          \\
Llama-2-13B     & meta-llama/Llama-2-13b-hf          \\
InternLM-20B    & internlm/internlm-20b              \\
Qwen-14B        & Qwen/Qwen-14B                      \\
Llama-2-7B      & meta-llama/Llama-2-7b-hf           \\
Qwen-7B         & Qwen/Qwen-7B                       \\
XGen-7B         & Salesforce/xgen-7b-8k-base         \\
Internlm-7B     & internlm/internlm-7b               \\
Phi-1\_5        & microsoft/phi-1\_5                 \\
Stablelm-7B     & stabilityai/stablelm-base-alpha-7b \\
Stablelm-3B     & stabilityai/stablelm-base-alpha-3b \\
Pythia-12B      & EleutherAI/pythia-12b              \\
Pythia-6.9B     & EleutherAI/pythia-6.9b             \\
Pythia-2.8B     & EleutherAI/pythia-2.8b             \\
Pythia-1B       & EleutherAI/pythia-1b               \\
Llama-2-70B     & meta-llama/Llama-2-70b-hf          \\
CodeLlama-34B   & codellama/CodeLlama-34b-hf         \\
CodeLlama-13B   & codellama/CodeLlama-13b-hf         \\
CodeLlama-7B    & codellama/CodeLlama-7b-hf          \\
StarCoder-15B   & bigcode/starcoderbase              \\
StarCoder-7B    & bigcode/starcoderbase-7b           \\
StarCoder-3B    & bigcode/starcoderbase-3b           \\
StarCoder-1B    & bigcode/starcoderbase-1b           \\
CodeGen-15B     & Salesforce/codegen-16B-multi       \\
CodeGen-6B      & Salesforce/codegen-6B-multi        \\
CodeGen-2B      & Salesforce/codegen-2B-multi        \\
Yarn-LLaMA2-13B & NousResearch/Yarn-Llama-2-7b-64k   \\
LongChat-13B    & lmsys/longchat-7b-16k              \\
RWKV-Raven-14B  & lmsys/longchat-7b-16k              \\ \bottomrule
\end{tabular}
\caption{LLMs used in our experiments and their corresponding names in Huggingface Hub.}
\label{tab:appendix_model}
\end{table*}

\subsection{Markdown vs. Flatten Setting Experiments}
\label{app:markdown_flatten}
{\small
\begin{verbatim}
"0": Size: 100 * 5, Template: Easy, Model: GPT-3.5
"1": Size: 50 * 5,  Template: Easy, Model: GPT-3.5
"2": Size: 20 * 6,  Template: Count, Model: GPT-3.5
"3": Size: 40 * 10, Template: Where Condition Text, Model: GPT-3.5
"4": Size: 10 * 20, Template: Where Condition Text, Model: GPT-3.5
"5": Size: 10 * 15, Template: Where Condition Text, Model: GPT-3.5
"6": Size: 50 * 5,  Template: Easy, Model: Llama-2-13B
"7": Size: 100 * 5, Template: Easy, Model: Yarn-Llama-2-13B
"8": Size: 50 * 5,  Template: Easy, Model: Yarn-Llama-2-13B
"9": Size: 25 * 7, Template: General, Model: Llama-2-13B
"10": Size: (15~40) * (6~9), Template: General, Model: Llama-2-13B
"11": Size: (15~40) * (6~9), Template: General, Model: Llama-2-13B 
"12": Size: (15~40) * (6~9), Template: Easy, Model: Llama-2-13B 
"13": Size: (15~40) * (6~9), Template: Easy, Model: Llama-2-13B
"14": Size: (15~40) * (6~9), Template: Easy, Model: Llama-2-13B
\end{verbatim}
}

\end{document}